\documentclass[runningheads]{llncs}

\usepackage{setspace}
\usepackage{eccv}

\usepackage{eccvabbrv}

\usepackage{graphicx}
\usepackage{booktabs}
\usepackage{multirow}
\usepackage{microtype}
\usepackage{enumitem}
\setlist[itemize]{leftmargin=*, noitemsep, topsep=2pt, parsep=0pt, partopsep=0pt}
\setlist[enumerate]{leftmargin=*, noitemsep, topsep=2pt, parsep=0pt, partopsep=0pt}

\usepackage[accsupp]{axessibility} 

\usepackage[pagebackref,breaklinks,colorlinks,citecolor=eccvblue]{hyperref}
\usepackage{hyperref}

\usepackage{orcidlink}

\begin{document}
\title{ReFP-AD: Rectified Flow Preconditioning  \\ for Energy-Based Anomaly Detection}

\titlerunning{ReFP-AD: Flow-Preconditioned EBMs for Anomaly Detection}

\author{Camile Lendering\ \and
Erkut Akdag \and
Joaquín Figueira \and
Egor Bondarev}

\authorrunning{C.~Lendering et al.}

\institute{
AIMS Group, Department of Electrical Engineering, Eindhoven University of Technology,
Eindhoven, The Netherlands\\
\email{c.r.lendering@tue.nl}
}

\maketitle
\begin{abstract}
Unified anomaly detection requires modeling highly heterogeneous normal data without access to anomalous samples. While foundation models like DINOv2 provide rich token representations, leveraging these spaces for explicit density estimation remains challenging. Energy-Based Models (EBMs) offer a principled formulation, but their training in high-dimensional token spaces is unstable due to anisotropy and strong cross-dimensional correlations, which degrades finite-step Markov Chain Monte Carlo (MCMC) sampling. 
We identify this instability as fundamentally geometric and introduce ReFP-AD (Rectified Flow Preconditioning for Anomaly Detection), which learns a geometric reparameterization that maps high-dimensional embeddings into a well-conditioned latent space via an optimal transport (OT)-coupled rectified flow. This preconditioning enables stable persistent contrastive divergence with preconditioned Stochastic Gradient Langevin Dynamics (SGLD) in full-dimensional token spaces. Anomaly scores are then derived from the learned energy landscape using gradient norms. Under a strict unified protocol on the MVTec-AD and VisA datasets, ReFP-AD achieves 98.6\%/97.9\% Image/Pixel AUROC on MVTec-AD and 97.3\%/99.0\% on VisA, outperforming prior unified EBM baselines by up to +10.8\% in Image AUROC. Ablation experiments demonstrate that geometric reparameterization is critical for finite-step MCMC and accurate anomaly localization in high-dimensional token spaces. Code is available at~\url{https://github.com/CLendering/ReFP-AD}
\end{abstract}

\section{Introduction}
\label{sec:intro}

Visual anomaly detection aims to identify inputs that deviate from the distribution of normal data without access to anomalous examples during training. This setting arises across industrial inspection, autonomous systems, and scientific imaging, where models must reliably detect rare and previously unseen deviations. In the~\emph{unified} anomaly detection setting, this challenge is further amplified: a single model must generalize across diverse object categories, textures, and appearance variations. Modeling such heterogeneous, yet exclusively normal visual distributions makes it difficult to learn stable decision boundaries in high-dimensional feature spaces.

Energy-based models (EBMs) offer a principled framework for modeling complex data distributions in high-dimensional spaces. Unlike likelihood-based generative models, they assign a scalar energy $E(x)$ to each input and define an unnormalized density $p(x) \propto \exp(-E(x))$~\cite{lecun2006tutorial}, enabling flexible representation of multi-modal distributions~\cite{du2019implicit} without restrictive architectural assumptions. For unsupervised anomaly detection, this formulation is compelling: anomalies are simply identified as high-energy inputs relative to the normal manifold.

In practice, EBM training requires sampling from the model distribution to approximate maximum-likelihood gradients, typically via Langevin dynamics~\cite{langevin1908theory} or related Markov chain Monte Carlo (MCMC) methods~\cite{nijkamp2019learning}. In high-dimensional spaces, these dynamics mix slowly and are sensitive to initialization, step size, and local curvature. As a result, practical EBM training relies on finite-step, non-equilibrium dynamics, including persistent contrastive divergence (PCD)~\cite{tieleman2008training}. Crucially, the stability of short-run sampling is determined not only by the energy function itself, but also by the geometry of the space in which sampling is performed.

In parallel, modern self-supervised Vision Transformers (ViTs) produce token embeddings that organize visual data into semantically structured manifolds~\cite{caron2021emerging,oquab2023dinov2}. These representations have proven highly effective for anomaly detection, often with simple similarity- or distance-based scoring~\cite{roth2022towards}. However, treating token embeddings as fixed feature spaces leaves explicit generative density modeling largely unexplored. While applying EBMs in token space is conceptually appealing, training with MCMC-based negative sampling remains sensitive and often unstable, especially under limited-step sampling and complex feature geometries~\cite{nijkamp2019learning,du2020improved}.

This limitation is fundamentally geometric rather than architectural. Standard Langevin dynamics corresponds to diffusion under a fixed Euclidean metric with isotropic noise~\cite{ma2015complete}. In contrast, transformer token distributions are highly anisotropic and strongly correlated across dimensions, violating the underlying Euclidean assumption~\cite{park2022how}. Under finite-step sampling, this metric mismatch yields poorly conditioned trajectories and unstable negative phases, necessitating delicate hyperparameter tuning. Consequently, prior EBMs often rely on compressing visual representations into low-dimensional bottlenecks~\cite{yoon2023energy}, sacrificing rich token-level semantics and limiting unified modeling capacity, to stabilize Markov chains.

To address this limitation, \emph{ReFP-AD} is proposed as a novel strategy to  explicitly reshape the sampling geometry~\emph{before} energy-based training. By leveraging the rectified flow formulation~\cite{liu2022flow}, our method learns an optimal transport (OT)-coupled map that transforms tokens into a well-conditioned latent coordinate system. Then, persistent contrastive divergence with preconditioned SGLD~\cite{li2016preconditioned} is performed in transported coordinates, effectively inducing a data-adaptive geometry under which the finite-step sampling becomes stable and predictable. Importantly, the energy model itself remains unconstrained; the intervention operates entirely at the level of representation geometry.

Rather than modifying the finite-step sampling procedure itself, the geometric intervention bridges the mismatch between high-dimensional semantic token spaces (e.g., 1536-D features for DINOv2-G) and finite-step MCMC. The rectified flow is optimized purely as a preconditioner, with model selection guided by finite-step Langevin diagnostics (conditioning, residual correlation, tail stability). This enables stable high-dimensional EBM training without dimensionality reduction.

Experiments on MVTec-AD and VisA demonstrate that the proposed framework achieves strong performance among density-based anomaly detectors. To summarize, the main contributions of this work are as follows:
\begin{itemize}
\item A geometric diagnosis of unified token-space EBM instability, where anisotropy and correlation in foundation-token representations violate the metric assumptions of standard Langevin dynamics.

\item \textbf{ReFP-AD}: a novel geometric reparameterization that maps high-dimensional vision tokens into an isotropic space conditioned for stable finite-step MCMC, enabling unconstrained EBMs in visual foundation-token spaces.

\item A comprehensive unified evaluation, demonstrating strong detection and localization performance (98.6/97.9\% Image/Pixel AUROC on MVTec-AD and 97.3/99.0\% on VisA), including targeted ablations isolating the impact of geometric conditioning under fixed sampling budgets.
\end{itemize}

\section{Related Work}
\label{sec:related_works}

\paragraph{Visual anomaly detection in pretrained feature spaces.}
Most modern industrial visual anomaly detection (VAD) methods avoid explicit generative modeling and, instead, score anomalies directly in pretrained feature spaces. Retrieval and distribution-based approaches like PatchCore~\cite{roth2022towards} and PaDiM~\cite{defard2021padim} remain strong baselines due to their reliable localization on benchmarks including MVTec-AD~\cite{bergmann2019mvtec}, VisA~\cite{zou2022spot}, and Real-IAD~\cite{wang2024real}. Subsequent work has emphasized practical deployment constraints, including real-time inference (EfficientAD~\cite{batzner2024efficientad}) and robustness to contaminated or noisy normal training data (SoftPatch~\cite{jiang2022softpatch}). However, these approaches rely on feature-space statistics rather than explicit probabilistic density modeling.

\paragraph{Foundation models and unified anomaly detection.}
Self-supervised Vision Transformers (ViTs) produce token embeddings with rich semantic and spatial structure~\cite{caron2021emerging,oquab2023dinov2}, further improving anomaly detection in feature-space. In few-shot regimes, simple token-level scoring on DINOv2 already yields competitive performance, whether via patch similarity (AnomalyDINO~\cite{damm2025anomalydino}) or subspace reconstruction residuals (SubspaceAD~\cite{subspacead}), indicating that pretrained tokens encode strong normality priors. Recent studies extend this paradigm to unified evaluation protocols spanning heterogeneous regimes, including semantic out-of-distribution (OOD) shifts and industrial defects (GeneralAD~\cite{strater2024generalad}). Vision-language models further expand this paradigm to category-scalable and zero-/few-shot settings through prompt alignment (WinCLIP~\cite{jeong2023winclip}, AnomalyCLIP~\cite{zhou2023anomalyclip}, PromptAD~\cite{li2024promptad}). Despite enabling open-set detection, they similarly default to discriminative scoring instead of explicit density estimation.

\paragraph{Density modeling in feature space.}
A complementary line of work models densities over intermediate representations using normalizing flows, thereby enabling tractable likelihood-based detection and localization. Methods, such as DifferNet~\cite{rudolph2021same}, FastFlow~\cite{yu2021fastflow}, and CFLOW-AD~\cite{gudovskiy2022cflow}, estimate feature-space likelihoods via invertible transformations, while diffusion-based approaches adopt reconstruction or denoising objectives for anomaly detection~\cite{he2024diffusion,zhang2025diffusionad,mousakhan2024anomaly}. More recent unified diffusion/flow-matching frameworks (e.g., DTG~\cite{wang2025debiasing}) introduce temporally adaptive guidance for multi-class anomaly detection under generative reconstruction settings. However, likelihood- and reconstruction-based models are known to assign high likelihoods or faithful reconstructions to OOD inputs~\cite{nalisnick2018deep,kirichenko2020normalizing}, highlighting sensitivity to representation geometry and objective design. In contrast, the proposed ReFP-AD model does not employ flows or diffusion models as generative estimators; instead a learned transport map is used as a geometric preconditioner to reshape the representation space prior to energy-based training. 

\paragraph{Energy-based models and non-equilibrium training.}
Energy-based models define unnormalized densities via scalar energy functions and are trained by lowering energy on data, while raising it on negative samples drawn from the model~\cite{lecun2006tutorial}. In practice, maximum-likelihood gradients are approximated by short-run Langevin dynamics, persistent contrastive divergence (PCD), or replay buffers. Prior analyses show that finite-step, non-equilibrium sampling choices, such as step size, number of updates, and initialization strategy, can qualitatively affect the learned energy landscape~\cite{nijkamp2020anatomy}. EBMs have also been applied to OOD detection through energy scores derived from discriminative models (Energy-OOD~\cite{liu2020energy}) and classifier-as-EBM formulations (JEM~\cite{grathwohl2019your}). Recent unified EBM approaches, including MPDR~\cite{yoon2023energy}, stabilize MCMC by collapsing visual representations into low-dimensional CNN feature vectors, trading representational richness for sampling stability. In contrast, the proposed ReFP-AD approach preserves full-dimensional foundation-token representations and addresses instability at the level of sampling geometry.

\paragraph{Transport maps and geometry-aware preconditioning.}
Sampling efficiency in high-dimensional correlated spaces is strongly influenced by geometry, motivating adaptive preconditioning and learned transport maps that transform targets into better-conditioned coordinates (e.g., NeuTra HMC~\cite{hoffman2019neutra}, latent-space Langevin for flows~\cite{nijkamp2020mcmc}, and EBM--flow hybrids~\cite{gao2020flow}). Rectified flow and flow matching provide stable ways to learn vector-field transports~\cite{liu2022flow,lipman2022flow}, and have recently been used directly for tabular anomaly detection via one-step contraction/deviation scoring~\cite{li2026scalable}. In contrast, ReFP-AD does not use flow matching as the detector: the rectified flow only preconditions high-dimensional visual tokens for finite-step PCD, while anomaly scores come from the learned EBM.

\section{Method}
\label{sec:method}

\begin{figure*}[ht]
 \centering
 \includegraphics[width=1.1\textwidth]{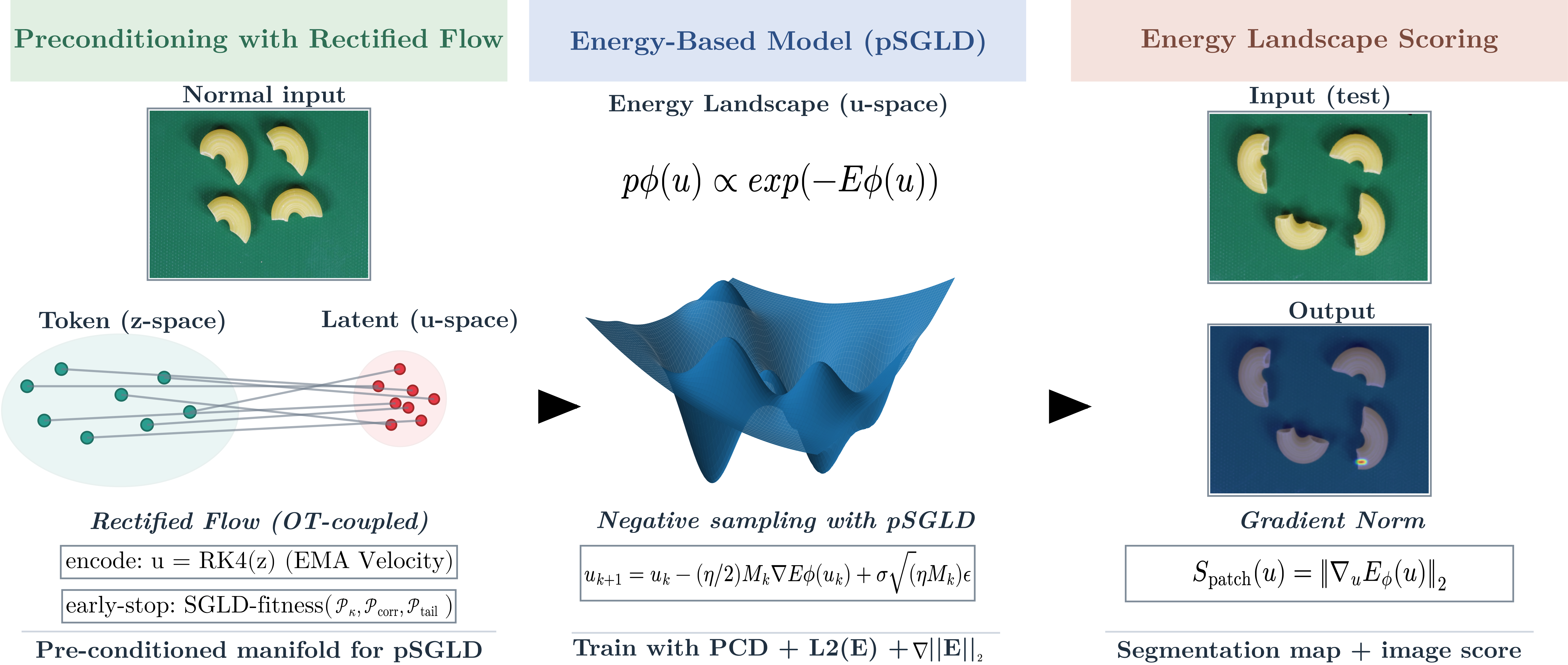}
 \caption{Overview of \textbf{ReFP-AD}. Standardized DINOv2 tokens in $z$ space are transported to a better-conditioned latent space $u$ via an OT-coupled rectified flow, with checkpoint selection guided by MCMC-oriented geometric diagnostics. An unconstrained EBM $E_\phi(u)$ is trained using PCD with pSGLD, and anomalies are scored at inference by the gradient norm $\|\nabla_u E_\phi(u)\|_2$ under a fixed finite-step sampling budget.}
 \label{fig:method_overview}
\end{figure*}

Figure~\ref{fig:method_overview} outlines our approach: standardized ViT tokens in $z$-space are first mapped into a well-conditioned latent $u$-space via OT-coupled rectified flow preconditioning, after which an unconstrained EBM is trained with PCD using pSGLD. This decouples representation geometry from density estimation: native token distributions are highly anisotropic and strongly correlated, so finite-step Langevin dynamics mixes poorly and destabilizes the negative phase. By learning a continuous-time transport map that reshapes tokens into approximately isotropic coordinates, both EBM training and gradient-norm anomaly scoring are performed entirely in the transported $u$-space, yielding stable MCMC behavior without architectural bottlenecks. 

\subsection{Unified Token Representation and Standardization}
\label{sec:tokens}

Given an input image, a frozen foundation ViT (DINOv2 ViT-G/14) produces a spatial grid of patch tokens $z_i \in \mathbb{R}^D$, where $D=1536$. To capture both low-level structure and high-level semantics, representations from multiple deep transformer layers are aggregated via mean pooling. Specifically, for each spatial patch index $i$, the corresponding tokens are averaged across a selected set of transformer layers $\mathcal{L}$, excluding the CLS token.

In the unified setting, token distributions across object categories can differ substantially in scale. Without normalization, these magnitude disparities dominate the multi-class optimization landscape and obscure the underlying geometric structure. To address this, tokens are Z-score standardized per category using normal training data:
\begin{equation}
    z \leftarrow \frac{z - \mu_c}{\sigma_c},
\end{equation}
where $\mu_c,\sigma_c \in \mathbb{R}^D$ are per-category feature mean and standard deviation estimated from normal training tokens, and the division is element-wise (with a small constant added to $\sigma_c$ for numerical stability).
This transformation constitutes a fixed affine pre-processing step. Subsequently, both the rectified flow and EBM are entirely unified, shared models that do not condition on category identity beyond this initial standardization.

\subsection{Geometric Preconditioning via Optimal Transport Flow}
\label{sec:flow}

Even after Z-score standardization, high-dimensional foundation-model token embeddings remain highly anisotropic, exhibiting strong cross-feature correlations and large covariance condition numbers. In such anisotropic spaces, MCMC literature establishes that finite-step Langevin dynamics produces inefficient, highly oscillatory ``zig-zag'' trajectories, where gradients dominate along narrow dimensions and mixing remains poor along elongated manifold directions~\cite{li2016preconditioned,hoffman2019neutra}. 

To eliminate this geometric limitation, a transport map is learned to reshape the token distribution into an isotropic geometry prior to EBM training. The target reference distribution is defined as an isotropic Gaussian $p_{\text{prior}}(u)=\mathcal{N}(0,\tau^2 I)$, where the temperature $\tau$ controls the scale of the latent space and the contraction–expansion balance of the transport, influencing conditioning and finite-step MCMC stability.

The transport is parameterized by a rectified flow~\cite{liu2022flow}, defined by the ordinary differential equation (ODE) $\dot{x}_t = v_\theta(t, x_t)$. To obtain geometrically consistent source–target pairings, data tokens $z$ and Gaussian targets $u \sim \mathcal{N}(0, \tau^2 I)$ are coupled via an entropic optimal transport (OT)-coupling $\pi(z, u)$. Since exact OT computations are numerically unstable in high dimensions ($1536$-D), the coupling $\pi$ is computed by stable log-domain Sinkhorn iterations~\cite{cuturi2013sinkhorn}. Given the linear interpolation $x_t = (1-t)z + t u$, the velocity network is optimized via the continuous-time matching objective:
\begin{equation}
 \mathcal{L}_{\text{RF}} = \mathbb{E}_{t \sim \mathcal{U}[0,1], (z, u) \sim \pi} \left[ \left\| v_\theta(t, x_t) - (u - z) \right\|_2^2 \right].
 \label{eq:rf_loss}
\end{equation}
Network weights are tracked by an Exponential Moving Average (EMA)~\cite{tarvainen2017mean} to ensure stable integration. At inference, tokens are transported to $u$ by integrating the ODE from t=0 to t=1 via a 4th-order Runge-Kutta (RK4) solver~\cite{roberts1996exponential}.

\subsection{SGLD-Oriented Manifold Validation}
\label{sec:flow_validation}

A low flow-matching loss $\mathcal{L}_{\text{RF}}$ ensures accurate vector-field regression, but does not guarantee improved~\emph{finite-step} Langevin behavior. Because the transport is introduced as a preconditioner for short-run PCD, flow checkpoints are selected using diagnostics that directly reflect known MCMC failure modes: ill-conditioning, residual correlation, and heavy-tailed instability.

Full covariance estimation in $D{=}1536$ is numerically unstable; therefore, transported tokens $u$ are projected onto a fixed random orthogonal subspace of dimension $k{=}128$, which preserves second-order structure in expectation while reducing estimator variance. Let $C$ denote the covariance of projected tokens. The following diagnostics are computed: (i) an anisotropy penalty $\mathcal{P}_{\kappa}=\log(\lambda_{\max}(C)/\lambda_{\min}(C))$ to reflect conditioning; (ii) a correlation penalty $\mathcal{P}_{\text{corr}}$, defined as the mean squared off-diagonal entries of the corresponding correlation matrix, to quantify residual cross-dimensional dependencies; and (iii) a tail penalty $\mathcal{P}_{\text{tail}}=\mathrm{q}_{0.99}(\|u\|_2)/\mathrm{median}(\|u\|_2)$ to control heavy-tailed outliers that destabilize SGLD updates.

These terms are combined into an SGLD-fitness score, represented by
\begin{equation}
    \mathcal{F}(u)=\mathcal{P}_{\kappa}+\tfrac{1}{2}\mathcal{P}_{\text{corr}}+\tfrac{1}{4}\log(\mathcal{P}_{\text{tail}}),
\end{equation}
where the coefficients are fixed globally across all experiments.
The weighting reflects the relative impact of each failure mode on short-run Langevin dynamics: conditioning typically dominates mixing behavior, residual correlations have secondary influence, and heavy-tail instability affects occasional but destabilizing updates.
The coefficients are selected once and kept constant; no dataset-specific tuning is performed.

To prevent degenerate manifold collapse, we enforce structural preservation via the Spearman rank correlation $\rho$ between pairwise distances in standardized space $z$ and transported space $u$. The optimal checkpoint is selected by solving $t^* = \arg\min_t \{ \mathcal{F}(u_t) \mid \rho_t \ge 0.6 \}$. This threshold ensures MCMC stability is prioritized only within a feasible region of topological integrity. All coefficients and the guardrail $\rho \ge 0.6$ are fixed across all datasets to avoid over-tuning. Visualizations are provided in Suppl. C.

\subsection{Unconstrained EBM with pSGLD}
\label{sec:ebm}

After transporting standardized tokens $z$ into the latent space $u$, density modeling is performed with a residual MLP energy function $E_\phi(u)$ with parameters $\phi$, defining the unnormalized density $p_\phi(u)\propto \exp(-E_\phi(u))$.

Training follows persistent contrastive divergence (PCD)~\cite{tieleman2008training}. To mitigate mode imbalance in the unified setting, the replay buffer employs vectorized stratified sampling across categories. Negative samples are updated using preconditioned Stochastic Gradient Langevin Dynamics (pSGLD)~\cite{li2016preconditioned}:
\begin{equation}
u_{k+1}=u_k-\frac{\eta}{2}M_k\nabla_u E_\phi(u_k)+\sigma\sqrt{\eta M_k}\,\xi_k,
\qquad \xi_k\sim\mathcal{N}(0,I),
\end{equation}
where $u_k$ denotes the $k$-th Langevin iterate, $\eta$ is the step size, $\sigma$ the noise scale, $\xi_k$ standard Gaussian noise, and $M_k$ a positive diagonal preconditioning matrix. 

The preconditioner is computed from an RMSProp-style moving average of squared gradients:
\begin{equation}
v_k=\beta v_{k-1}+(1-\beta)\big(\nabla_u E_\phi(u_k)\big)^2,
\qquad
M_k=(\sqrt{v_k}+\epsilon)^{-1},
\end{equation}
where $v_k$ is the running second-moment estimate, $\beta\in[0,1)$ the decay factor, and $\epsilon>0$ a small constant for numerical stability.

The EBM is optimized with the contrastive objective
\begin{equation}
\mathcal{L}_{\text{EBM}} =
\mathbb{E}_{u^+}[E_\phi(u)] - \mathbb{E}_{u^-}[E_\phi(u)]
+ \alpha \mathbb{E}_{u^\pm}[E_\phi(u)^2]
+ \lambda \mathbb{E}_{u^+}[\|\nabla_u E_\phi(u)\|_2^2],
\label{eq:ebm_loss}
\end{equation}
where $u^+$ and $u^-$ denote positive samples from the data distribution and negative samples from the replay buffer, respectively. $\mathbb{E}$ denotes expectation over the corresponding sample sets. The coefficients $\alpha$ and $\lambda$ control energy magnitude regularization and gradient smoothness.

\subsection{Energy Landscape Scoring}
\label{sec:scoring}

During inference, normal samples concentrate near low-energy minima, whereas anomalies occupy higher-energy states. Therefore, anomaly scoring is derived directly from the learned energy landscape, by the energy gradient norm as a local deviation measure:
\begin{equation}
S_{\text{patch}}(u) = \left\|\nabla_u E_\phi(u)\right\|_2,
\label{eq:patch_score}
\end{equation}
which computes the magnitude of the “restoring force” required to move a sample toward the learned normal manifold~\cite{grathwohl2019your}. Due to the well-conditioned latent space induced by ReFP-AD, the raw energy gradient norm provides a stable and informative signal for out-of-distribution deviation, reducing reliance on explicit priors or iterative refinement heuristics.

The obtained scores are projected onto the 2D token grid, bilinearly upsampled, and smoothed with a Gaussian filter ($\sigma$=4.0). To ensure robustness to localized noise spikes, the image-level anomaly score is computed as the mean of the top $1\%$ of the upsampled pixel scores~\cite{damm2025anomalydino}. While this gradient-based score forms our primary detection signal,  a complementary ablation (Suppl.~A) shows that the transported token magnitude $\|u\|_2$ already provides a strong baseline, but explicit energy modeling improves robustness on challenging categories.

\section{Experiments and Results}
\label{sec:experiments}

This section evaluates~\emph{ReFP-AD} under the strict unified protocol on MVTec-AD and VisA. Main results, ablations, and qualitative localization are reported. Additional unified results on MVTec-AD 2 and Real-IAD are reported in Suppl.~D.

\subsection{Experimental Setup}
\label{sec:exp_setup}

\paragraph{Datasets and the Unified Evaluation Protocol.}
The proposed ReFP-AD method is evaluated on two industrial anomaly detection benchmarks: MVTec-AD~\cite{bergmann2019mvtec} (15 categories) and VisA~\cite{zou2022spot} (12 categories). A strict~\emph{unified} setting is adopted, where a~\emph{single shared} rectified flow and a~\emph{single shared} EBM are trained jointly on all normal images across categories, without per-category networks, heads, or separate training runs. Category identity is limited to fixed input standardization (Sec.~\ref{sec:tokens}) and for balanced replay-buffer initialization, reflecting industrial scenarios where category labels are available at test time.

\paragraph{Implementation Details.}
Input images are resized to $700 \times 700$ and center-cropped to $672 \times 672$. Patch tokens (size $14 \times 14$) are extracted by a frozen DINOv2 ViT-G/14 backbone, aggregating representations from the middle seven transformer layers (layers $22$ to $28$) via mean pooling to obtain $1536$-dimensional feature vectors. The rectified flow is parameterized by an 8-layer MLP with hidden dimension $1024$. It is trained for up to $150$ epochs with batch size $8192$ via the Adam optimizer and a learning rate of $5\times 10^{-5}$. The log-domain Sinkhorn OT-coupling employs an entropic regularization of $0.01$, and the target Gaussian noise temperature is set to $\tau=0.1$. To ensure integration stability, an Exponential Moving Average (EMA) with decay $0.999$ is applied to the velocity network weights. Flow optimization utilizes dynamic early stopping based on the SGLD-fitness criterion with a patience of $20$ epochs. At inference stage, flow integration exploits a 10-step 4th-order Runge-Kutta (RK4) solver. 

The unconstrained EBM is a 3-layer residual MLP with a hidden dimension of $1024$. Training is performed for $15$ epochs with batch size $8192$ using Adam (learning rate $10^{-5}$, weight decay $10^{-5}$, with $\gamma=0.4$ decay at epochs 8 and 12). The objective is stabilized with an $L_2$ energy penalty ($\alpha=0.1$) and a gradient norm penalty ($\lambda=10.0$). The PCD replay buffer contains $400,000$ states and employs a $5\%$ stratified re-initialization rate from the data manifold to mitigate multi-class mode collapse. Negative sampling executes $60$ steps of preconditioned SGLD per update with step size of $10^{-3}$, momentum $\beta=0.99$, and noise standard deviation $0.1$.

\subsection{Unified Anomaly Detection Performance}
\label{sec:main_results}

While recent unified detectors often rely on similarity or reconstruction scoring in pretrained feature spaces, this work focuses on density-based modeling with EBMs; therefore, our primary comparisons are restricted to explicit density estimators, with a contextual comparison to recent reconstruction-based methods provided in Suppl.~B. For context, the evaluated methods are grouped into general baseline approaches, normalizing-flow-based methods, and energy-based models.

\begin{table}[ht]
\centering
\caption{Unified anomaly detection and localization performance on MVTec-AD and VisA, reported as Image AUROC / Pixel AUROC (\%). Baseline results are from HGAD~\cite{yao2024hierarchical} and stabilized MPDR$^{\dagger}$~\cite{yoon2023energy}.}
\label{tab:unified_main}
\begin{tabular*}{\textwidth}{@{\extracolsep{\fill}} l cccc}
\toprule
\multirow{2}{*}{Method} & \multicolumn{2}{c}{MVTec-AD} & \multicolumn{2}{c}{VisA} \\
\cmidrule(lr){2-3} \cmidrule(lr){4-5}
& I-AUROC & P-AUROC & I-AUROC & P-AUROC \\
\midrule
\multicolumn{5}{l}{\textit{Baseline Methods}} \\
\midrule
PaDiM \cite{defard2021padim} & 84.2 & 89.5 & 86.8 & 97.0 \\
MKD \cite{salehi2021multiresolution} & 81.9 & 84.9 & 74.2 & 93.9 \\
DRAEM \cite{zavrtanik2021draem} & 88.1 & 87.2 & 85.5 & 90.5 \\
\midrule
\multicolumn{5}{l}{\textit{Normalizing Flow Based Methods}} \\
\midrule
FastFlow \cite{yu2021fastflow} & 91.8 & 96.0 & 77.2 & 95.1 \\
CFLOW \cite{gudovskiy2022cflow} & 89.0 & 94.0 & 88.0 & 95.9 \\
HGAD \cite{yao2024hierarchical} & \underline{98.4} & \textbf{97.9} & \underline{97.1} & \underline{98.9} \\
\midrule
\multicolumn{5}{l}{\textit{Energy-Based Models}} \\
\midrule
EBM (Genc et al.)~\cite{genc2021energy} & 72.0 & 70.7 & - & - \\
MPDR$^{\dagger}$ \cite{yoon2023energy} & 96.0 & 96.7 & 86.5 & 96.5 \\
ReFP-AD (Ours) & \textbf{98.6} & \textbf{97.9} & \textbf{97.3} & \textbf{99.0} \\
\bottomrule
\end{tabular*}
\end{table}

The proposed ReFP-AD model achieves the strongest performance among the methods considered in Table~\ref{tab:unified_main}, with particularly large gains over prior EBM baselines in the unified VisA setting. Notably, it outperforms the leading normalizing-flow-based model, HGAD~\cite{yao2024hierarchical}, achieving $98.6\%$ Image AUROC on MVTec-AD and $97.3\%$ on the heterogeneous VisA dataset. These results indicate that, when the representation space is properly conditioned, explicit density estimation can surpass hierarchical likelihood formulations in a unified multi-class setting.

A substantial margin is also observed over prior EBM approaches that stabilize training through recovery-based objectives anchored to autoencoder manifolds. MPDR~\cite{yoon2023energy} serves as a representative baseline of this paradigm,  employing a recovery-likelihood formulation in which sampling is regularized by a reconstruction-fidelity term that constrains MCMC near a low-dimensional autoencoder manifold (272-D CNN features in the unified setting). In the unified VisA reproduction with the official MPDR implementation, the default SGLD configuration was numerically unstable and resulted in near-random performance. To avoid underestimating the baseline, a stabilized MPDR variant ($\dagger$) is reported, obtained by tuning only SGLD hyperparameters to ensure convergence, yielding $86.5\%$ Image AUROC.

ReFP-AD improves upon the tuned MPDR baseline by $+10.8\%$ on VisA, highlighting the importance of geometric conditioning for stable high-dimensional EBM training. 
This improvement suggests that, in unified high-resolution settings, stability obtained through manifold-anchored recovery objectives alone is insufficient when the underlying representation geometry remains highly anisotropic. By directly conditioning the representation geometry, ReFP-AD enables stable sampling without architectural compression or reconstruction-based energy terms, while preserving the structured token information required for accurate localization and multi-class modeling.

\subsection{Ablation Studies}
\label{sec:ablation}

To validate the core components of the proposed method, ablation studies are conducted on the MVTec-AD and VisA datasets. These experiments isolate the impact of geometric preconditioning, unified vs. per-category modeling, the number of SGLD sampling steps, backbone scale, and sampling preconditioning. Additional robustness analyses for DINOv3-7B features and input resolution are provided in Suppl.~G and Suppl.~F.

\begin{figure}[t]
 \centering
 \includegraphics[width=0.8\textwidth]{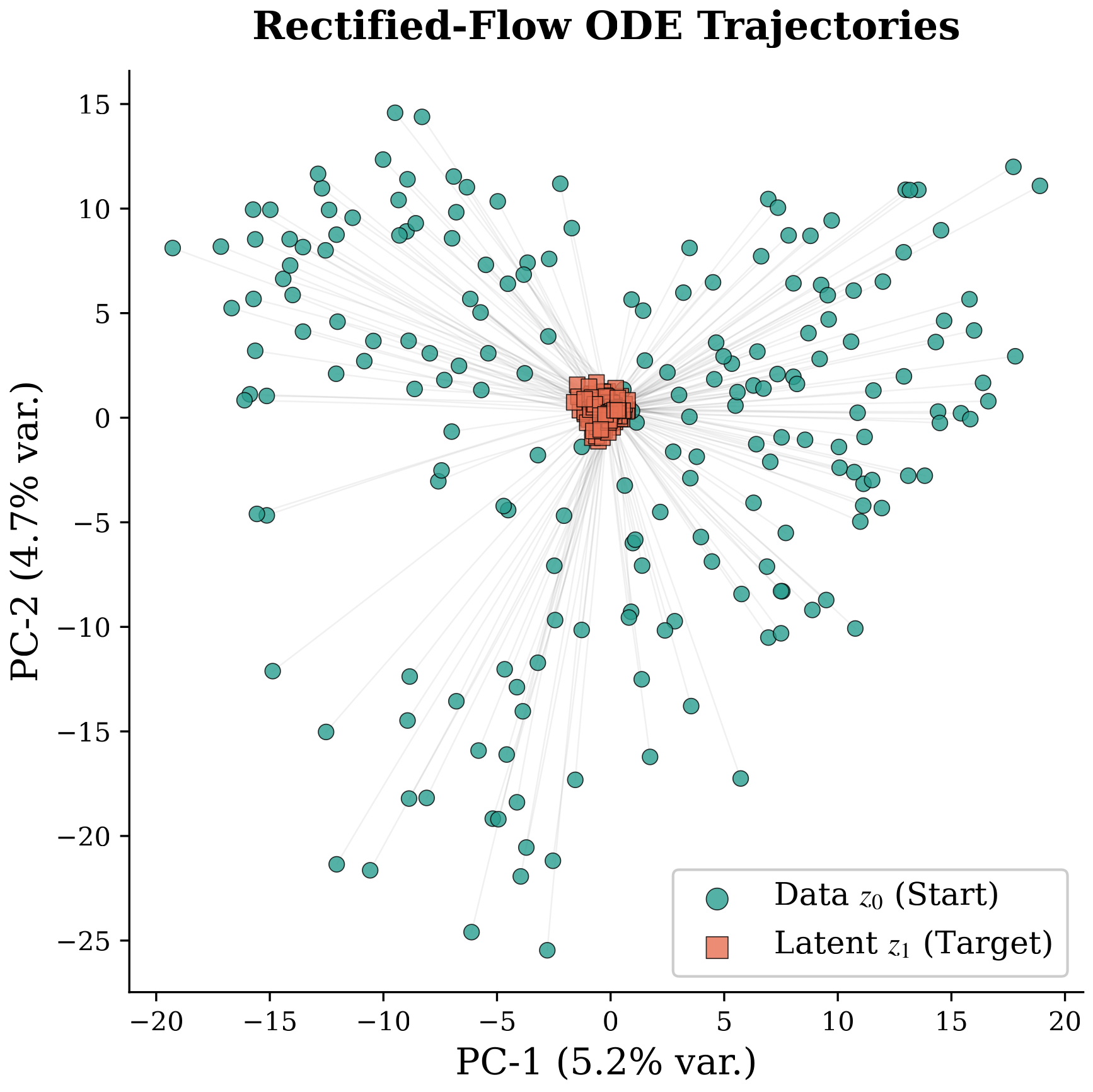}
 \caption{PCA projection of rectified flow trajectories. Axes correspond to the first two principal components (PC-1: 5.2\% variance, PC-2: 4.7\% variance). The transport map transforms the correlated, heavy-tailed token distribution ($z$) into a well-conditioned latent space ($u$) with reduced anisotropy and correlations, improving finite-step Langevin dynamics.}
 \label{fig:flow_trajectories}

\end{figure}

\subsubsection{Effect of Rectified Flow Preconditioning.}
The fundamental hypothesis of this work is that raw foundation-token spaces are geometrically unsuitable for finite-step Langevin dynamics. To evaluate this, the unconstrained EBM is trained directly on the standardized $1536$-dimensional DINOv2 token space, without optimal transport preconditioning. Removing the geometric preconditioning yields a substantial degradation in performance, particularly on the heterogeneous VisA dataset, where the image-level AUROC drops from $97.3\%$ down to $87.0\%$ ($\Delta -10.3\%$) and pixel-level AUROC decreases to $96.6\%$ ($\Delta -2.4\%$). On the comparatively simpler MVTec-AD benchmark, detection performance also declines significantly from $98.6\%$ to $91.1\%$ ($\Delta -7.5\%$). Figure~\ref{fig:flow_trajectories} illustrates how the rectified flow reshapes, decorrelates, and isotropizes the empirical token manifold. 
In the raw anisotropic space, finite-step Langevin dynamics mixes poorly, leading to unstable negative-phase updates and collapsed energy basins. These results demonstrate that geometric preconditioning is essential for stable EBM training in high-dimensional dense foundation-token representations.

\subsubsection{Unified vs. Per-Category Modeling.}
Conventional industrial anomaly detection models are typically trained per-category, operating under a simplified, unimodal optimization landscape. Table~\ref{tab:ablation_routing} compares the proposed unified model (ReFP-AD) against its per-category variant, as well as against SimpleNet~\cite{liu2023simplenet}, a competitive per-category baseline.

\begin{table}[ht]

\centering
\caption{Comparison of Per-Category vs. Unified modeling. Performance is reported as Image AUROC (\%). The proposed unified model remains highly competitive with dedicated per-category models, exhibiting only marginal performance degradation despite the increased complexity of modeling all categories simultaneously.}
\label{tab:ablation_routing}
\begin{tabular}{l c @{\hspace{2.5em}} c c}
\toprule
Method & Setting & MVTec-AD & VisA \\
\midrule
SimpleNet \cite{liu2023simplenet} & Per-Category & 99.6 & 96.9 \\
ReFP-AD & Per-Category & 99.2 & 97.7 \\
\midrule
ReFP-AD & Unified & 98.6 & 97.3 \\
\midrule
\textit{Unified $\Delta$ (vs. Per-Cat)} & - & \textbf{-0.6} & \textbf{-0.4} \\
\bottomrule
\end{tabular}
\end{table}

The results indicate a marginal performance gap between the dedicated per-category networks and the unified model. Specifically, the unified model trails its per-category counterpart by only $0.6\%$ on MVTec-AD and $0.4\%$ on the more complex VisA dataset. In the context of generative anomaly detection, where forcing a single model to capture $12$ to $15$ distinct semantic manifolds, often increasing the risk of mode imbalance and degraded density boundaries, this minimal degradation is notable.

\subsubsection{Effect of pSGLD Sampling Steps.}
Energy-based models are known to be highly sensitive to the number of Langevin steps applied during the negative phase, often requiring hundreds of iterations to mix properly in high-dimensional spaces~\cite{nijkamp2019learning}. However, the proposed ReFP-AD maps the data into an isotropic, well-conditioned latent space, thereby enabling rapid mixing under preconditioned SGLD.

\begin{table}[ht]

\centering
\caption{Ablation on the number of preconditioned SGLD steps ($K$) during the EBM negative phase on the unified MVTec-AD benchmark. The flow-preconditioned manifold enables rapid MCMC mixing, stabilizing detection performance in as few as 20 steps.}
\label{tab:ablation_sgld}
\begin{tabular}{l @{\hspace{3em}} c @{\hspace{2em}} c @{\hspace{2em}} c @{\hspace{2em}} c}
\toprule
SGLD Steps ($K$) & 10 & 20 & 40 & 80 \\
\midrule
Image AUROC (\%) & 54.7 & 98.2 & 98.5 & 98.6 \\
Pixel AUROC (\%) & 67.8 & 94.1 & 97.9 & 97.9 \\
\bottomrule
\end{tabular}

\end{table}

As shown in Table~\ref{tab:ablation_sgld}, ablating the number of SGLD steps ($K \in \{10, 20, 40, 80\}$) on the unified MVTec-AD benchmark reveals rapid performance saturation. With only $K=10$ steps, the Markov chains fail to mix sufficiently, resulting in a collapsed image AUROC of $54.7\%$. However, increasing it to $K=20$ steps leads to a substantial improvement, with image-level performance rising to $98.2\%$. By $K=40$ steps, pixel-level localization fully stabilizes ($98.5\%$ I-AUROC, $97.9\%$ P-AUROC), while further increasing the chains to $K=80$ steps yields negligible deviations ($98.6\%$ I-AUROC, $97.9\%$ P-AUROC). These results demonstrate that geometric preconditioning substantially improves the sampling conditioning, so stable negative-phase updates can be obtained with relatively few finite-step pSGLD iterations.

\subsubsection{Backbone Scalability.}
To analyze the robustness of the ReFP-AD model to backbone capacity, we evaluate its performance under different DINOv2 variants. DINOv2 ViT-G/14 is replaced with ViT-L/14 and ViT-B/14, while all other components are kept fixed. As shown in Table~\ref{tab:ablation_backbone}, performance degrades only moderately with reduced backbone size, indicating that the proposed preconditioning and EBM training remain stable across token spaces ranging from 768D--1536D.

\begin{table}[ht]
\centering
\caption{Ablation of backbone scale on unified Image AUROC and Pixel AUROC. The proposed ReFP-AD maintains stable performance across feature dimensions of 768D, 1024D, and 1536D.}
\label{tab:ablation_backbone}
\setlength{\tabcolsep}{0pt}
\begin{tabular*}{\textwidth}{@{\extracolsep{\fill}} l c cc cc}
\toprule
\multirow{2}{*}{Backbone} & \multirow{2}{*}{Dimension} & \multicolumn{2}{c}{MVTec-AD} & \multicolumn{2}{c}{VisA} \\
\cmidrule(lr){3-4} \cmidrule(lr){5-6}
& & I-AUROC & P-AUROC & I-AUROC & P-AUROC \\
\midrule
DINOv2 ViT-B/14 & 768  & 97.7 & 97.6 & 95.7 & 98.8 \\
DINOv2 ViT-L/14 & 1024 & 97.6 & 97.6 & 96.7 & 98.9 \\
DINOv2 ViT-G/14 & 1536 & \textbf{98.6} & \textbf{97.9} & \textbf{97.3} & \textbf{99.0} \\
\bottomrule
\end{tabular*}

\end{table}

\subsubsection{Necessity of SGLD Preconditioning.}
Table~\ref{tab:ablation_precond} compares preconditioned SGLD (pSGLD)~\cite{li2016preconditioned} with standard SGLD on the VisA dataset. Removing the diagonal preconditioner leads to a significant drop in image AUROC (97.3$\rightarrow$75.3), indicating that while transport improves global conditioning, adaptive local scaling is still required for stable negative-phase mixing in the learned energy landscape. Both samplers are tuned over the same step-size/noise search range under the unified VisA setting.

\begin{table}[ht]

\centering
\caption{Ablation of the sampling dynamics on the unified VisA benchmark. Comparing standard SGLD with the proposed preconditioned SGLD (pSGLD). The adaptive preconditioner is essential for maintaining stable detection performance across complex, multi-class energy landscapes.}
\label{tab:ablation_precond}
\begin{tabular}{l c c}
\toprule
Sampling Method & I-AUROC & P-AUROC \\
\midrule
Standard SGLD & 75.3 & 94.3 \\
Preconditioned SGLD (pSGLD) & \textbf{97.3} & \textbf{99.0} \\
\bottomrule
\end{tabular}

\end{table}

\subsection{Qualitative Analysis}
\label{sec:qualitative}
\begin{figure}[ht!]
 \centering
 \includegraphics[width=\textwidth]{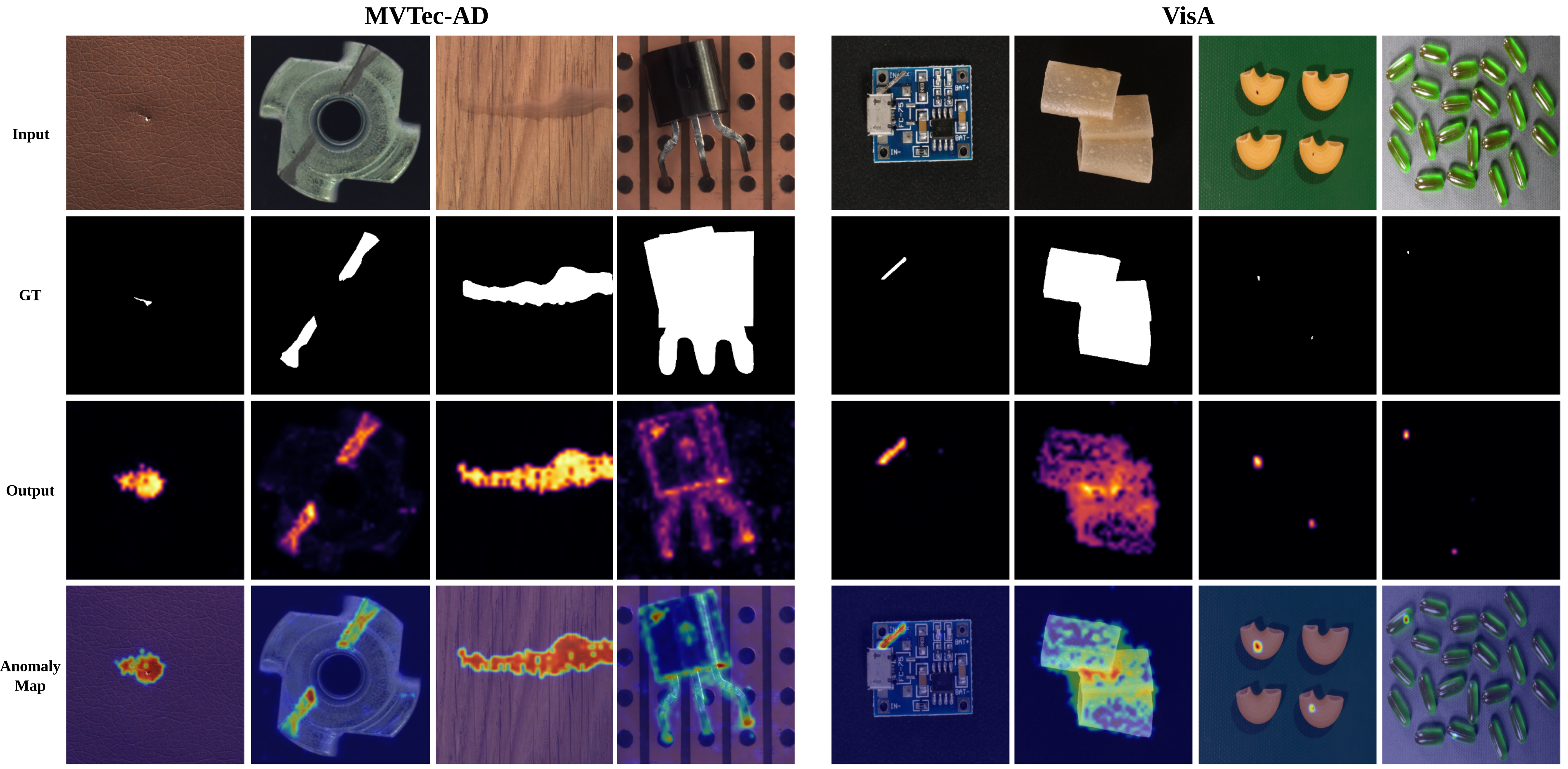}
 \caption{Qualitative localization results on MVTec-AD and VisA datasets. From top to bottom: input image, ground-truth mask, predicted anomaly map, and overlay. The proposed model ReFP-AD produces well-localized anomaly maps that closely align with defect regions, while maintaining low scores on normal backgrounds.}
 \label{fig:mvtec_visa}

\end{figure}

Figure~\ref{fig:mvtec_visa} (and extended results in Suppl.~H) shows anomaly maps obtained via gradient-norm scoring. Operating on dense transported tokens yields sharp localization of both subtle texture and structural defects, while suppressing background noise by concentrating normal regions near low-energy basins.

\subsection{Limitations and Computational Overhead}
\label{sec:limitations}

While ReFP-AD stabilizes energy-based density modeling in high-dimensional token spaces, it increases inference cost. In contrast to feature-based detectors, such as SimpleNet~\cite{liu2023simplenet} and PatchCore~\cite{roth2022towards}, which operate primarily in a feedforward manner or rely on efficient k-NN retrieval, the proposed method transports tokens with a 10-step RK4 solver and computes the gradient-norm score $S_{\text{patch}}(u)=\|\nabla_u E_\phi(u)\|_2$ via a backward pass. Although both the rectified flow and energy networks are lightweight MLPs, the combination of ODE integration and gradient evaluation leads to increased latency. Consequently, the method is better suited to offline inspection or settings where unified modeling outweighs strict real-time constraints. Latency benchmarks are reported in Suppl.~E.

\section{Conclusion}
This work shows that the key limitation in training EBMs on modern foundation-token representations is not model capacity, but~\emph{sampling geometry}. In high-dimensional ViT token spaces, strong anisotropy and cross-dimensional correlations violate the assumptions underlying finite-step Langevin dynamics, making PCD unstable and forcing prior unified EBMs to rely on architectural bottlenecks or recovery constraints. \emph{ReFP-AD} addresses this limitation through a geometric reparameterization: an OT-coupled rectified flow transports standardized tokens into a well-conditioned latent space in which short-run MCMC becomes stable. This enables~\emph{unconstrained} energy learning directly on dense high-dimensional DINOv2 token representations under a strict unified protocol, yielding strong detection and localization performance on the MVTec-AD and VisA datasets. In particular, ReFP-AD achieves $98.6\%/97.9\%$ Image/Pixel AUROC on MVTec-AD and $97.3\%/99.0\%$ on VisA, improving unified EBM baselines by up to $+10.8\%$ Image AUROC.

Beyond these results, ReFP-AD suggests a general principle for scaling energy-based modeling to foundation representations:~\emph{first condition the representation geometry, then learn the energy}. Future work includes distilling the preconditioned energy landscape into an efficient feed-forward scorer to bridge the gap between stable unified modeling and latency-critical applications.

\section*{Acknowledgement}
This work is supported by the ADVISOR ITEA 241007 project.

%
%
\bibliographystyle{splncs04}
\bibliography{main}

@String(AAAI  = {AAAI})

@String(ICIP  = {IEEE Int. Conf. Image Process.})

@String(ICIP  = {ICIP})

@inproceedings{defard2021padim,
  title={Padim: a patch distribution modeling framework for anomaly detection and localization},
  author={Defard, Thomas and Setkov, Aleksandr and Loesch, Angelique and Audigier, Romaric},
  booktitle={International conference on pattern recognition},
  pages={475--489},
  year={2021},
  organization={Springer}
}

@article{yoon2023energy,
  title={Energy-based models for anomaly detection: A manifold diffusion recovery approach},
  author={Yoon, Sangwoong and Jin, Young-Uk and Noh, Yung-Kyun and Park, Frank},
  journal={Advances in Neural Information Processing Systems},
  volume={36},
  pages={49445--49466},
  year={2023}
}

@inproceedings{yao2024hierarchical,
  title={Hierarchical gaussian mixture normalizing flow modeling for unified anomaly detection},
  author={Yao, Xincheng and Li, Ruoqi and Qian, Zefeng and Wang, Lu and Zhang, Chongyang},
  booktitle={European Conference on Computer Vision},
  pages={92--108},
  year={2024},
  organization={Springer}
}

@inproceedings{gudovskiy2022cflow,
  title={Cflow-ad: Real-time unsupervised anomaly detection with localization via conditional normalizing flows},
  author={Gudovskiy, Denis and Ishizaka, Shun and Kozuka, Kazuki},
  booktitle={Proceedings of the IEEE/CVF winter conference on applications of computer vision},
  pages={98--107},
  year={2022}
}

@article{yu2021fastflow,
  title={Fastflow: Unsupervised anomaly detection and localization via 2d normalizing flows},
  author={Yu, Jiawei and Zheng, Ye and Wang, Xiang and Li, Wei and Wu, Yushuang and Zhao, Rui and Wu, Liwei},
  journal={arXiv preprint arXiv:2111.07677},
  year={2021}
}

@article{you2022unified,
  title={A unified model for multi-class anomaly detection},
  author={You, Zhiyuan and Cui, Lei and Shen, Yujun and Yang, Kai and Lu, Xin and Zheng, Yu and Le, Xinyi},
  journal={Advances in Neural Information Processing Systems},
  volume={35},
  pages={4571--4584},
  year={2022}
}

@inproceedings{salehi2021multiresolution,
  title={Multiresolution knowledge distillation for anomaly detection},
  author={Salehi, Mohammadreza and Sadjadi, Niousha and Baselizadeh, Soroosh and Rohban, Mohammad H and Rabiee, Hamid R},
  booktitle={Proceedings of the IEEE/CVF conference on computer vision and pattern recognition},
  pages={14902--14912},
  year={2021}
}

@inproceedings{zavrtanik2021draem,
  title={Draem-a discriminatively trained reconstruction embedding for surface anomaly detection},
  author={Zavrtanik, Vitjan and Kristan, Matej and Sko{\v{c}}aj, Danijel},
  booktitle={Proceedings of the IEEE/CVF international conference on computer vision},
  pages={8330--8339},
  year={2021}
}

@inproceedings{roth2022towards,
  title={Towards total recall in industrial anomaly detection},
  author={Roth, Karsten and Pemula, Latha and Zepeda, Joaquin and Sch{\"o}lkopf, Bernhard and Brox, Thomas and Gehler, Peter},
  booktitle={Proceedings of the IEEE/CVF conference on computer vision and pattern recognition},
  pages={14318--14328},
  year={2022}
}

@inproceedings{bergmann2019mvtec,
  title={MVTec AD--A comprehensive real-world dataset for unsupervised anomaly detection},
  author={Bergmann, Paul and Fauser, Michael and Sattlegger, David and Steger, Carsten},
  booktitle={Proceedings of the IEEE/CVF conference on computer vision and pattern recognition},
  pages={9592--9600},
  year={2019}
}

@inproceedings{zou2022spot,
  title={Spot-the-difference self-supervised pre-training for anomaly detection and segmentation},
  author={Zou, Yang and Jeong, Jongheon and Pemula, Latha and Zhang, Dongqing and Dabeer, Onkar},
  booktitle={European conference on computer vision},
  pages={392--408},
  year={2022},
  organization={Springer}
}

@inproceedings{wang2024real,
  title={Real-iad: A real-world multi-view dataset for benchmarking versatile industrial anomaly detection},
  author={Wang, Chengjie and Zhu, Wenbing and Gao, Bin-Bin and Gan, Zhenye and Zhang, Jiangning and Gu, Zhihao and Qian, Shuguang and Chen, Mingang and Ma, Lizhuang},
  booktitle={Proceedings of the IEEE/CVF Conference on Computer Vision and Pattern Recognition},
  pages={22883--22892},
  year={2024}
}

@inproceedings{batzner2024efficientad,
  title={Efficientad: Accurate visual anomaly detection at millisecond-level latencies},
  author={Batzner, Kilian and Heckler, Lars and K{\"o}nig, Rebecca},
  booktitle={Proceedings of the IEEE/CVF winter conference on applications of computer vision},
  pages={128--138},
  year={2024}
}

@article{jiang2022softpatch,
  title={Softpatch: Unsupervised anomaly detection with noisy data},
  author={Jiang, Xi and Liu, Jianlin and Wang, Jinbao and Nie, Qiang and Wu, Kai and Liu, Yong and Wang, Chengjie and Zheng, Feng},
  journal={Advances in Neural Information Processing Systems},
  volume={35},
  pages={15433--15445},
  year={2022}
}

@inproceedings{caron2021emerging,
  title={Emerging properties in self-supervised vision transformers},
  author={Caron, Mathilde and Touvron, Hugo and Misra, Ishan and J{\'e}gou, Herv{\'e} and Mairal, Julien and Bojanowski, Piotr and Joulin, Armand},
  booktitle={Proceedings of the IEEE/CVF international conference on computer vision},
  pages={9650--9660},
  year={2021}
}

@article{oquab2023dinov2,
  title={Dinov2: Learning robust visual features without supervision},
  author={Oquab, Maxime and Darcet, Timoth{\'e}e and Moutakanni, Th{\'e}o and Vo, Huy and Szafraniec, Marc and Khalidov, Vasil and Fernandez, Pierre and Haziza, Daniel and Massa, Francisco and El-Nouby, Alaaeldin and others},
  journal={arXiv preprint arXiv:2304.07193},
  year={2023}
}

@inproceedings{strater2024generalad,
  title={Generalad: Anomaly detection across domains by attending to distorted features},
  author={Str{\"a}ter, Luc PJ and Salehi, Mohammadreza and Gavves, Efstratios and Snoek, Cees GM and Asano, Yuki M},
  booktitle={European conference on computer vision},
  pages={448--465},
  year={2024},
  organization={Springer}
}

@inproceedings{jeong2023winclip,
  title={Winclip: Zero-/few-shot anomaly classification and segmentation},
  author={Jeong, Jongheon and Zou, Yang and Kim, Taewan and Zhang, Dongqing and Ravichandran, Avinash and Dabeer, Onkar},
  booktitle={Proceedings of the IEEE/CVF conference on computer vision and pattern recognition},
  pages={19606--19616},
  year={2023}
}

@article{zhou2023anomalyclip,
  title={Anomalyclip: Object-agnostic prompt learning for zero-shot anomaly detection},
  author={Zhou, Qihang and Pang, Guansong and Tian, Yu and He, Shibo and Chen, Jiming},
  journal={arXiv preprint arXiv:2310.18961},
  year={2023}
}

@inproceedings{li2024promptad,
  title={Promptad: Learning prompts with only normal samples for few-shot anomaly detection},
  author={Li, Xiaofan and Zhang, Zhizhong and Tan, Xin and Chen, Chengwei and Qu, Yanyun and Xie, Yuan and Ma, Lizhuang},
  booktitle={Proceedings of the IEEE/CVF Conference on Computer Vision and Pattern Recognition},
  pages={16838--16848},
  year={2024}
}

@inproceedings{damm2025anomalydino,
  title={Anomalydino: Boosting patch-based few-shot anomaly detection with dinov2},
  author={Damm, Simon and Laszkiewicz, Mike and Lederer, Johannes and Fischer, Asja},
  booktitle={2025 IEEE/CVF Winter Conference on Applications of Computer Vision (WACV)},
  pages={1319--1329},
  year={2025},
  organization={IEEE}
}

@inproceedings{rudolph2021same,
  title={Same same but differnet: Semi-supervised defect detection with normalizing flows},
  author={Rudolph, Marco and Wandt, Bastian and Rosenhahn, Bodo},
  booktitle={Proceedings of the IEEE/CVF winter conference on applications of computer vision},
  pages={1907--1916},
  year={2021}
}

@inproceedings{he2024diffusion,
  title={A diffusion-based framework for multi-class anomaly detection},
  author={He, Haoyang and Zhang, Jiangning and Chen, Hongxu and Chen, Xuhai and Li, Zhishan and Chen, Xu and Wang, Yabiao and Wang, Chengjie and Xie, Lei},
  booktitle={Proceedings of the AAAI conference on artificial intelligence},
  volume={38},
  pages={8472--8480},
  year={2024}
}

@article{zhang2025diffusionad,
  title={DiffusionAD: Norm-guided one-step denoising diffusion for anomaly detection},
  author={Zhang, Hui and Wang, Zheng and Zeng, Dan and Wu, Zuxuan and Jiang, Yu-Gang},
  journal={IEEE transactions on pattern analysis and machine intelligence},
  year={2025},
  publisher={IEEE}
}

@inproceedings{mousakhan2024anomaly,
  title={Anomaly detection with conditioned denoising diffusion models},
  author={Mousakhan, Arian and Brox, Thomas and Tayyub, Jawad},
  booktitle={DAGM German Conference on Pattern Recognition},
  pages={181--195},
  year={2024},
  organization={Springer}
}

@article{nalisnick2018deep,
  title={Do deep generative models know what they don't know?},
  author={Nalisnick, Eric and Matsukawa, Akihiro and Teh, Yee Whye and Gorur, Dilan and Lakshminarayanan, Balaji},
  journal={arXiv preprint arXiv:1810.09136},
  year={2018}
}

@article{kirichenko2020normalizing,
  title={Why normalizing flows fail to detect out-of-distribution data},
  author={Kirichenko, Polina and Izmailov, Pavel and Wilson, Andrew G},
  journal={Advances in neural information processing systems},
  volume={33},
  pages={20578--20589},
  year={2020}
}

@article{liu2020energy,
  title={Energy-based out-of-distribution detection},
  author={Liu, Weitang and Wang, Xiaoyun and Owens, John and Li, Yixuan},
  journal={Advances in neural information processing systems},
  volume={33},
  pages={21464--21475},
  year={2020}
}

@article{grathwohl2019your,
  title={Your classifier is secretly an energy based model and you should treat it like one},
  author={Grathwohl, Will and Wang, Kuan-Chieh and Jacobsen, J{\"o}rn-Henrik and Duvenaud, David and Norouzi, Mohammad and Swersky, Kevin},
  journal={arXiv preprint arXiv:1912.03263},
  year={2019}
}

@article{hoffman2019neutra,
  title={Neutra-lizing bad geometry in hamiltonian monte carlo using neural transport},
  author={Hoffman, Matthew and Sountsov, Pavel and Dillon, Joshua V and Langmore, Ian and Tran, Dustin and Vasudevan, Srinivas},
  journal={arXiv preprint arXiv:1903.03704},
  year={2019}
}

@article{nijkamp2020mcmc,
  title={Mcmc should mix: Learning energy-based model with neural transport latent space mcmc},
  author={Nijkamp, Erik and Gao, Ruiqi and Sountsov, Pavel and Vasudevan, Srinivas and Pang, Bo and Zhu, Song-Chun and Wu, Ying Nian},
  journal={arXiv preprint arXiv:2006.06897},
  year={2020}
}

@inproceedings{gao2020flow,
  title={Flow contrastive estimation of energy-based models},
  author={Gao, Ruiqi and Nijkamp, Erik and Kingma, Diederik P and Xu, Zhen and Dai, Andrew M and Wu, Ying Nian},
  booktitle={Proceedings of the IEEE/CVF Conference on Computer Vision and Pattern Recognition},
  pages={7518--7528},
  year={2020}
}

@article{liu2022flow,
  title={Flow straight and fast: Learning to generate and transfer data with rectified flow},
  author={Liu, Xingchao and Gong, Chengyue and Liu, Qiang},
  journal={arXiv preprint arXiv:2209.03003},
  year={2022}
}

@article{lipman2022flow,
  title={Flow matching for generative modeling},
  author={Lipman, Yaron and Chen, Ricky TQ and Ben-Hamu, Heli and Nickel, Maximilian and Le, Matt},
  journal={arXiv preprint arXiv:2210.02747},
  year={2022}
}

@article{cuturi2013sinkhorn,
  title={Sinkhorn distances: Lightspeed computation of optimal transport},
  author={Cuturi, Marco},
  journal={Advances in neural information processing systems},
  volume={26},
  year={2013}
}

@inproceedings{li2016preconditioned,
  title={Preconditioned stochastic gradient Langevin dynamics for deep neural networks},
  author={Li, Chunyuan and Chen, Changyou and Carlson, David and Carin, Lawrence},
  booktitle={Proceedings of the AAAI conference on artificial intelligence},
  volume={30},
  year={2016}
}

@article{lecun2006tutorial,
  title={A tutorial on energy-based learning},
  author={LeCun, Yann and Chopra, Sumit and Hadsell, Raia and Ranzato, M and Huang, Fujie and others},
  journal={Predicting structured data},
  volume={1},
  number={0},
  year={2006}
}

@article{du2019implicit,
  title={Implicit generation and modeling with energy based models},
  author={Du, Yilun and Mordatch, Igor},
  journal={Advances in neural information processing systems},
  volume={32},
  year={2019}
}

@article{nijkamp2019learning,
  title={Learning non-convergent non-persistent short-run mcmc toward energy-based model},
  author={Nijkamp, Erik and Hill, Mitch and Zhu, Song-Chun and Wu, Ying Nian},
  journal={Advances in Neural Information Processing Systems},
  volume={32},
  year={2019}
}

@inproceedings{tieleman2008training,
  title={Training restricted Boltzmann machines using approximations to the likelihood gradient},
  author={Tieleman, Tijmen},
  booktitle={Proceedings of the 25th international conference on Machine learning},
  pages={1064--1071},
  year={2008}
}

@article{genc2021energy,
  title={Energy-based anomaly detection and localization},
  author={Genc, Ergin Utku and Ahuja, Nilesh and Ndiour, Ibrahima J and Tickoo, Omesh},
  journal={arXiv preprint arXiv:2105.03270},
  year={2021}
}

@article{ma2015complete,
  title={A complete recipe for stochastic gradient MCMC},
  author={Ma, Yi-An and Chen, Tianqi and Fox, Emily},
  journal={Advances in neural information processing systems},
  volume={28},
  year={2015}
}

@inproceedings{liu2023simplenet,
  title={Simplenet: A simple network for image anomaly detection and localization},
  author={Liu, Zhikang and Zhou, Yiming and Xu, Yuansheng and Wang, Zilei},
  booktitle={Proceedings of the IEEE/CVF conference on computer vision and pattern recognition},
  pages={20402--20411},
  year={2023}
}

@inproceedings{dinomaly,
  title={Dinomaly: The less is more philosophy in multi-class unsupervised anomaly detection},
  author={Guo, Jia and Lu, Shuai and Zhang, Weihang and Chen, Fang and Li, Huiqi and Liao, Hongen},
  booktitle={Proceedings of the Computer Vision and Pattern Recognition Conference},
  pages={20405--20415},
  year={2025}
}

@inproceedings{wang2025debiasing,
  title={Debiasing Trace Guidance: Top-down Trace Distillation and Bottom-up Velocity Alignment for Unsupervised Anomaly Detection},
  author={Wang, Xingjian and Chai, Li and Chen, Jiming},
  booktitle={Proceedings of the IEEE/CVF International Conference on Computer Vision},
  pages={22989--22998},
  year={2025}
}

@article{du2020improved,
  title={Improved contrastive divergence training of energy based models},
  author={Du, Yilun and Li, Shuang and Tenenbaum, Joshua and Mordatch, Igor},
  journal={arXiv preprint arXiv:2012.01316},
  year={2020}
}

@article{tarvainen2017mean,
  title={Mean teachers are better role models: Weight-averaged consistency targets improve semi-supervised deep learning results},
  author={Tarvainen, Antti and Valpola, Harri},
  journal={Advances in neural information processing systems},
  volume={30},
  year={2017}
}

@article{langevin1908theory,
  title={On the theory of brownian motion.},
  author={Langevin, Paul},
  journal={CR Acad Sci (Paris)},
  volume={146},
  pages={530},
  year={1908}
}

@inproceedings{akcay2022anomalib,
  title={Anomalib: A deep learning library for anomaly detection},
  author={Akcay, Samet and Ameln, Dick and Vaidya, Ashwin and Lakshmanan, Barath and Ahuja, Nilesh and Genc, Utku},
  booktitle={2022 IEEE International Conference on Image Processing (ICIP)},
  pages={1706--1710},
  year={2022},
  organization={IEEE}
}

@article{heckler2025mvtec,
  title={The mvtec ad 2 dataset: Advanced scenarios for unsupervised anomaly detection},
  author={Heckler-Kram, Lars and Neudeck, Jan-Hendrik and Scheler, Ulla and K{\"o}nig, Rebecca and Steger, Carsten},
  journal={arXiv preprint arXiv:2503.21622},
  year={2025}
}

@inproceedings{
park2022how,
title={How Do Vision Transformers Work?},
author={Namuk Park and Songkuk Kim},
booktitle={International Conference on Learning Representations},
year={2022},
url={https://openreview.net/forum?id=D78Go4hVcxO}
}

@inproceedings{nijkamp2020anatomy,
  title={On the anatomy of mcmc-based maximum likelihood learning of energy-based models},
  author={Nijkamp, Erik and Hill, Mitch and Han, Tian and Zhu, Song-Chun and Wu, Ying Nian},
  booktitle={Proceedings of the AAAI Conference on Artificial Intelligence},
  volume={34},
  pages={5272--5280},
  year={2020}
}

@article{roberts1996exponential,
  author    = {Roberts, Gareth O. and Tweedie, Richard L.},
  title     = {Exponential convergence of {L}angevin distributions and their discrete approximations},
  journal   = {Bernoulli},
  volume    = {2},
  number    = {4},
  pages     = {341--363},
  year      = {1996},
  month     = {December},
  publisher = {Bernoulli Society for Mathematical Statistics and Probability},
  doi       = {10.2307/3318418},
  URL       = {https://projecteuclid.org/journals/bernoulli/volume-2/issue-4/Exponential-convergence-of-Langevin-distributions-and-their-discrete-approximations/10.2307/3318418.full}
}

@inproceedings{deng2022anomaly,
  title={Anomaly detection via reverse distillation from one-class embedding},
  author={Deng, Hanqiu and Li, Xingyu},
  booktitle={Proceedings of the IEEE/CVF conference on computer vision and pattern recognition},
  pages={9737--9746},
  year={2022}
}

@article{he2024mambaad,
  title={Mambaad: Exploring state space models for multi-class unsupervised anomaly detection},
  author={He, Haoyang and Bai, Yuhu and Zhang, Jiangning and He, Qingdong and Chen, Hongxu and Gan, Zhenye and Wang, Chengjie and Li, Xiangtai and Tian, Guanzhong and Xie, Lei},
  journal={Advances in Neural Information Processing Systems},
  volume={37},
  pages={71162--71187},
  year={2024}
}

@article{guo2023recontrast,
  title={Recontrast: Domain-specific anomaly detection via contrastive reconstruction},
  author={Guo, Jia and Lu, Shuai and Jia, Lize and Zhang, Weihang and Li, Huiqi},
  journal={Advances in Neural Information Processing Systems},
  volume={36},
  pages={10721--10740},
  year={2023}
}

@article{zhang2023exploring,
  title={Exploring plain vit reconstruction for multi-class unsupervised anomaly detection},
  author={Zhang, Jiangning and Chen, Xuhai and Wang, Yabiao and Wang, Chengjie and Liu, Yong and Li, Xiangtai and Yang, Ming-Hsuan and Tao, Dacheng},
  journal={arXiv preprint arXiv:2312.07495},
  year={2023}
}

@article{luo2025inp,
  title={Inp-former++: Advancing universal anomaly detection via intrinsic normal prototypes and residual learning},
  author={Luo, Wei and Yao, Haiming and Cao, Yunkang and Chen, Qiyu and Gao, Ang and Shen, Weiming and Yu, Wenyong},
  journal={arXiv preprint arXiv:2506.03660},
  year={2025}
}

@article{li2026scalable,
  title={Scalable, explainable and provably robust anomaly detection with one-step flow matching},
  author={Li, Zhong and Huang, Qi and Zhu, Yuxuan and Yang, Lincen and Mohammadi Amiri, Mohammad and van Stein, Niki and van Leeuwen, Matthijs},
  journal={Advances in Neural Information Processing Systems},
  volume={38},
  pages={87834--87899},
  year={2026}
}

@inproceedings{subspacead,
  title={SubspaceAD: Training-Free Few-Shot Anomaly Detection via Subspace Modeling},
  author={Lendering, Camile and Akdag, Erkut and Bondarau, Egor},
  booktitle={Proceedings of the IEEE/CVF Conference on Computer Vision and Pattern Recognition},
  pages={28557--28566},
  year={2026}
}

\clearpage
\appendix
\setcounter{section}{0}
\title{Supplementary Material for ReFP-AD: Rectified Flow Preconditioning  \\ for Energy-Based Anomaly Detection} 

\titlerunning{Rectified Flow Preconditioning for Anomaly Detection}

\titlerunning{ReFP-AD: Flow-Preconditioned EBMs for Anomaly Detection}

\author{Camile Lendering\ \and
Erkut Akdag \and
Joaquín Figueira \and
Egor Bondarev}

\authorrunning{C.~Lendering et al.}

\institute{
AIMS Group, Department of Electrical Engineering, Eindhoven University of Technology,
Eindhoven, The Netherlands\\
\email{c.r.lendering@tue.nl}
}

\maketitle

\setcounter{table}{0}
\setcounter{figure}{0}
\renewcommand{\thetable}{\Alph{section}.\arabic{table}}
\renewcommand{\thefigure}{\Alph{section}.\arabic{figure}}

\section{Flow-Native Anomaly Scores vs.\ Energy Learning}
\label{sec:app_flow_scores}

\paragraph{Motivation.}
ReFP-AD reshapes the geometry of foundation-model token distributions to stabilize finite-step MCMC during EBM training (Sec.~3.2--3.4).
However, the learned transport may also induce a~\emph{direct} anomaly signal without explicit energy modeling, for instance through the magnitude of transported tokens.
This section investigates such \emph{flow-native} scoring and clarifies the complementary role of the EBM objective.

\paragraph{Flow-magnitude scoring.}
Given transported tokens $u \in \mathbb{R}^{D}$, a purely geometric patch score is defined as
\begin{equation}
S_{\text{mag}}(u) = \|u\|_2,
\label{eq:app_flow_mag}
\end{equation}
computed independently per token.
Patch scores are projected back to the image plane using the same interpolation and Gaussian smoothing procedure as in Sec.~3.5.
The image-level score is computed as the mean of the top $1\%$ pixel scores.

\paragraph{Comparison protocol.}
The magnitude-based score $S_{\text{mag}}$ is compared with the primary EBM score $S_{\text{patch}}(u)=\|\nabla_u E_\phi(u)\|_2$ under the identical unified evaluation protocol and identical post-processing.
This comparison isolates whether explicit energy learning provides benefits beyond the geometry induced by transport.

\paragraph{Results and interpretation.}
Table~\ref{tab:app_flow_vs_ebm} compares pure geometric magnitude scoring
$S_{\text{mag}}(u)=\|u\|_2$ with the learned EBM gradient-norm score.
The transported magnitude already yields strong unified performance
(96.18\% I-AUROC on VisA and 97.51\% on MVTec-AD), indicating that
the learned rectified flow alone induces a meaningful anomaly signal.

Explicit energy learning further improves average performance and, more importantly, enhances worst-case robustness.
On the VisA dataset, the most challenging category (\texttt{macaroni2}) improves
from 85.69\% to 93.06\% I-AUROC (+7.37\%).
On MVTec-AD, the hardest category (\texttt{screw}) improves from
82.84\% to 89.92\% I-AUROC (+7.08\%), and \texttt{capsule} improves
from 89.43\% to 95.85\% (+6.42\%). These gains are concentrated in the most challenging categories, where magnitude-based separation alone is insufficient to achieve high I-AUROC.
This observation suggests that geometric conditioning provides a strong baseline, while energy learning primarily enhances robustness under unified multi-class heterogeneity.

\begin{table}[t]
\centering
\caption{Flow-magnitude vs.\ EBM scoring under unified training.
Average and worst-category Image AUROC (\%).}
\label{tab:app_flow_vs_ebm}
\begin{tabular}{l @{\hspace{2em}} cccc}
\toprule
\multicolumn{5}{c}{\textbf{MVTec-AD}} \\
\midrule
\multirow{2}{*}{Scoring} & Avg & Worst & Hardest & \multirow{2}{*}{$\Delta$ Worst} \\
 & I-AUROC & I-AUROC & Cat. & \\
\midrule
Flow magnitude ($\|u\|_2$) & 97.51 & 82.84 & screw & -- \\
EBM grad-norm (Ours) & \textbf{98.60} & \textbf{89.92} & screw & \textbf{+7.08} \\
\midrule
\multicolumn{5}{c}{\textbf{VisA}} \\
\midrule
Flow magnitude ($\|u\|_2$) & 96.18 & 85.69 & macaroni2 & -- \\
EBM grad-norm (Ours) & \textbf{97.32} & \textbf{93.06} & macaroni2 & \textbf{+7.37} \\
\bottomrule
\end{tabular}
\end{table}

\section{Comparison with Reconstruction-Based Methods}
\label{sec:app_reconstruction_sota}

\paragraph{Motivation.}
The main paper focuses on density- and energy-based anomaly detection, where the central question is whether explicit energy modeling can be made effective in high-dimensional foundation-token spaces.
However, many recent state-of-the-art unified anomaly detection methods are reconstruction- or decoder-based.
We therefore provide a comparison with strong reconstruction-based methods to clarify both the added value and the remaining gap of the proposed EBM formulation.

\begin{table}[ht]
\centering
\caption{
Comparison with reconstruction- and decoder-based multi-class UAD methods.
We report Image AUROC / Pixel AUROC (\%).
ReFP-AD is not reconstruction-based; it is included to clarify the added value and remaining gap of the proposed EBM formulation.
}
\label{tab:app_reconstruction_sota}
\setlength{\tabcolsep}{4pt}
\small
\begin{tabular*}{\textwidth}{@{\extracolsep{\fill}} l l cc}
\toprule
Method & Modeling principle & MVTec-AD & VisA \\
\midrule
RD4AD~\cite{deng2022anomaly}
& Reverse distillation
& 94.6 / 96.1
& 92.4 / 98.1 \\

UniAD~\cite{you2022unified}
& Transformer reconstruction
& 96.5 / 96.8
& 88.8 / 98.3 \\

ReContrast~\cite{guo2023recontrast}
& Reconstruction contrast
& 98.3 / 97.1
& 95.5 / 98.5 \\

ViTAD~\cite{zhang2023exploring}
& ViT reconstruction
& 98.3 / 97.7
& 90.5 / 98.2 \\

MambaAD~\cite{he2024mambaad}
& State-space reconstruction
& 98.6 / 97.7
& 94.3 / 98.5 \\

Dinomaly~\cite{dinomaly}
& Feature reconstruction
& 99.7 / 98.4
& \textbf{98.9} / 98.8 \\

INP-Former++~\cite{luo2025inp}
& INP-guided residual decoding
& \textbf{99.8} / \textbf{98.7}
& \textbf{98.9} / \textbf{99.1} \\

\midrule
ReFP-AD (Ours)
& Flow-preconditioned EBM
& 98.6 / 97.9
& 97.3 / 99.0 \\
\bottomrule
\end{tabular*}
\end{table}

\paragraph{Discussion.}
Table~\ref{tab:app_reconstruction_sota} shows that the proposed EBM formulation does not fully close the gap to the strongest reconstruction- and decoder-based methods in image-level AUROC.
This is expected, since methods such as Dinomaly and INP-Former++ train explicit feature reconstruction or residual-decoding objectives whose outputs are directly aligned with dense anomaly localization.
These models therefore benefit from task-specific decoder structures and reconstruction losses that encourage spatially resolved residual maps.

In contrast, ReFP-AD does not train a reconstruction decoder, does not use a segmentation head, and does not define anomalies through reconstruction residuals.
Instead, it learns an explicit energy landscape over full-dimensional foundation-model token embeddings.
The rectified flow is used only as a geometric preconditioner to make finite-step MCMC stable in this high-dimensional token space.
Therefore, the purpose of ReFP-AD is not to replace highly optimized reconstruction decoders, but to demonstrate that explicit density/energy modeling becomes competitive once the representation geometry is properly conditioned.

The comparison highlights this distinction.
Although ReFP-AD remains below the strongest reconstruction SOTA in image-level detection, it reaches a similar localization regime, achieving 97.9\% Pixel AUROC on MVTec-AD and 99.0\% on VisA without relying on reconstruction fidelity as the anomaly signal.
This supports the added value of the proposed formulation: ReFP-AD isolates geometric preconditioning as a mechanism for stable token-space EBM training and provides an explicit energy-based anomaly score rather than a decoder residual.

\section{Flow Checkpoint Selection via SGLD Diagnostics}
\label{sec:app_flow_selection}

\paragraph{Motivation.}
The rectified flow is used as a geometric preconditioner for finite-step Langevin dynamics, rather than as a standalone generative model.
Therefore, minimizing the flow-matching objective $\mathcal{L}_{\mathrm{RF}}$ alone does not guarantee optimal downstream anomaly detection performance. Instead, checkpoint selection is guided by the introduced SGLD-fitness criterion, which directly measures geometric suitability for short-run MCMC.

\begin{figure}[t]
\centering
\includegraphics[width=\textwidth]{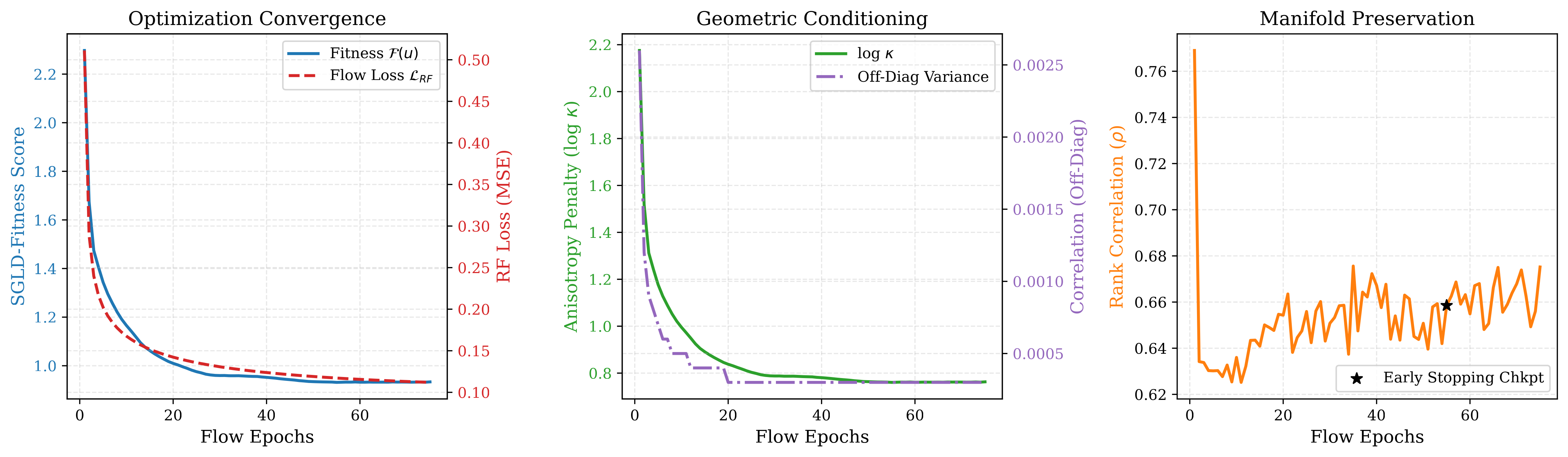}
\caption{
Flow training diagnostics.
Left: SGLD-fitness score $\mathcal{F}(u)$ (blue) and RF loss (red).
Middle: anisotropy penalty and off-diagonal covariance.
Right: Spearman rank correlation for manifold preservation. While the RF loss decreases monotonically, $\mathcal{F}(u)$ exhibits a clear optimum, which is used for early stopping.
}
\label{fig:flow_diagnostics}
\end{figure}

\paragraph{Convergence behavior.}
Figure~\ref{fig:flow_diagnostics} illustrates the evolution of the RF loss, the SGLD-fitness score $\mathcal{F}(u)$, and the associated geometric diagnostics.
While the RF loss decreases monotonically during training, the SGLD-fitness score exhibits a clear interior minimum.
Beyond this point, further improvements in vector-field regression do not translate into better conditioning for finite-step Langevin sampling.

\paragraph{Empirical validation.}
To verify this behavior quantitatively, the rectified flow is trained for 250 epochs and the final checkpoint is selected, based on minimizing $\mathcal{L}_{\mathrm{RF}}$, without MCMC-based model selection.
Under this setting (unified VisA training with DINOv2-G features), the average performance reaches
\[
\text{I-AUROC} = 73.05\%,
\qquad
\text{P-AUROC} = 94.77\%.
\]
While pixel-level localization remains strong, image-level discrimination degrades substantially for several structurally challenging categories (e.g., \textit{pcb3}, \textit{pcb1}, \textit{macaroni1}), indicating that excessive manifold conditioning does not improve short-run MCMC separability.

In contrast, selecting the checkpoint at the minimum of $\mathcal{F}(u)$ consistently produces higher image-level AUROC while maintaining strong localization performance.
This confirms that regression optimality of the transport map is not aligned with optimal conditioning for finite-step Langevin dynamics.

\paragraph{Interpretation.}
Early flow training reduces anisotropy and cross-dimensional correlation, improving the local geometry of the transported manifold.
However, excessive optimization of $\mathcal{L}_{\mathrm{RF}}$ produces diminishing geometric benefits and can distort relative distance structure important for density contrast.
The SGLD-fitness criterion balances improving conditioning with manifold preservation, leading to superior downstream EBM performance.

These observations support the central claim of this work:
transport quality for anomaly detection is determined not by regression accuracy alone, but by its effect on short-run MCMC behavior.

\section{Additional Datasets}
\label{sec:supl_additional_datasets}

\paragraph{Motivation.}
Recent benchmarks have been introduced to address the saturation of earlier industrial anomaly detection datasets. In particular, Real-IAD~\cite{wang2024real} expands the diversity of object categories and defect types under more realistic acquisition conditions, while MVTec-AD 2~\cite{heckler2025mvtec} provides higher-resolution imagery and more precise pixel-level annotations. Therefore, these datasets pose substantially greater challenges in terms of multi-class heterogeneity, fine-grained localization, and domain variability.

Evaluating ReFP-AD on these benchmarks serves two purposes. First, it examines whether geometric preconditioning remains effective under increased dataset complexity.
Second, given the recency of these datasets and the limited availability of published unified baselines, it provides a reference point for flow-based and energy-based methods under consistent training protocols.

\paragraph{Evaluation Protocol.}

All methods are evaluated under the same unified training regime as in the main paper, where a single shared model is jointly trained across all categories. To ensure a fair comparison, all baselines are trained with comparable computational budgets and identical image preprocessing. Specifically, images are resized and center-cropped to match the resolution used by ReFP-AD, and segmentation masks are evaluated at the same spatial resolution across methods.

Whenever available, official implementations are employed. Otherwise, baselines are reproduced within the \textsc{Anomalib} framework~\cite{akcay2022anomalib}.

For the Real-IAD dataset, evaluation follows the standard protocol, reporting Image-AUROC and Pixel-AUROC. For the MVTec-AD 2 dataset, image-level performance is measured using Image-AUROC, while segmentation quality is assessed via AU-PRO$_{0.05}$~\cite{bergmann2019mvtec}, consistent with the benchmark specification~\cite{heckler2025mvtec}. All MVTec-AD 2 results are reported on the official public test split.

The only deviations from the main-paper configuration concern sampling depth and model capacity. For both MPDR and our method ReFP-AD, the number of SGLD steps is increased from 60 to 150 to improve mixing under the larger-scale and more heterogeneous setting. In addition, the hidden dimensionality of both the flow and EBM MLPs is increased to 1536 to better accommodate the substantially larger dataset. All remaining hyperparameters are kept identical.

\paragraph{Results and Interpretation.}

\begin{table}[t]
\centering
\caption{Unified anomaly detection and localization performance on Real-IAD and MVTec-AD 2 datasets, reported as Image-AUROC / Pixel-AUROC (\%) for Real-IAD and Image-AUROC / AU-PRO$_{0.05}$ (\%) for MVTec-AD 2. Results are obtained using the original code-base when available, and otherwise reproduced within the Anomalib~\cite{akcay2022anomalib} framework.}
\label{tab:unified_suppl}
\setlength{\tabcolsep}{0pt} 
\begin{tabular*}{\textwidth}{@{\extracolsep{\fill}} l cccc}
\toprule
\multirow{2}{*}{Method} & \multicolumn{2}{c}{Real-IAD} & \multicolumn{2}{c}{MVTec-AD 2} \\
\cmidrule(lr){2-3} \cmidrule(lr){4-5}
& I-AUROC & P-AUROC & I-AUROC & AUPRO \\
\midrule
\multicolumn{5}{l}{\textit{Normalizing Flow Based Methods}} \\
\midrule
FastFlow \cite{yu2021fastflow} & 53.2 & 76.5 & 52.3 & 10.6 \\
CFLOW \cite{gudovskiy2022cflow} & 61.4 & 93.0 & 63.0 & 23.8 \\
HGAD \cite{yao2024hierarchical} & 82.4 & 92.1 & 62.0 & 18.4 \\
\midrule
\multicolumn{5}{l}{\textit{Energy-Based Models}} \\
\midrule
MPDR$^{\dagger}$ \cite{yoon2023energy} & 71.1 & 95.3 & 67.1 & 12.9 \\
ReFP-AD (Ours) & \textbf{82.5} & \textbf{98.8} & \textbf{72.4} & \textbf{38.6} \\
\bottomrule
\end{tabular*}
\end{table}

Table~\ref{tab:unified_suppl} summarizes unified performance on the Real-IAD and MVTec-AD 2 datasets.
As expected, the increased dataset complexity results in a performance drop across all evaluated methods, particularly in image-level discrimination.
Several flow-based approaches exhibit near-random I-AUROC on Real-IAD, indicating instability under large-scale multi-class heterogeneity. In contrast, ReFP-AD maintains competitive image-level performance while substantially improving localization robustness.

On the Real-IAD dataset, ReFP-AD achieves the highest Pixel-AUROC (98.8\%) and the strongest image-level performance among the evaluated methods (82.5\% I-AUROC).
On the MVTec-AD 2 dataset, ReFP-AD outperforms competing flow and energy-based methods by a notable margin in both Image-AUROC and AU-PRO$_{0.05}$, demonstrating that geometric preconditioning remains effective even at finer defect granularity and increased dataset complexity.

These results reinforce the central claim of this work: stabilizing the sampling geometry enables scalable and robust unified density modeling, particularly in regimes where dataset heterogeneity and resolution amplify the limitations of conventional flow-based or recovery-based objectives.

\section{Computational Complexity}
\label{sec:appendix_overhead}

\paragraph{Motivation and Setup.}
As discussed in Section~4.5 of the main paper, the proposed method ReFP-AD prioritizes stable, unified density modeling over real-time inference. To provide a transparent breakdown of the computational overhead, the unified inference pipeline is profiled on a single NVIDIA H100 GPU. The DINOv2-G/14 backbone is kept strictly frozen; therefore, its parameters are excluded from the trainable footprint.

\paragraph{Parameter and Memory Footprint.}
Despite operating in a 1536-dimensional token space, the trainable components remain lightweight. The model introduces exactly 13.12M trainable parameters, comprising 9.45M for the Rectified Flow network and 3.67M for the EBM. During inference, the peak VRAM footprint is measured at 5198.5 MB.

\paragraph{Latency and Throughput.}
With cached DINOv2-G/14 features, inference requires a Runge-Kutta 4th Order (RK4) ODE integration for geometric transport and a backward pass through the EBM to compute the gradient-norm anomaly score. Under this configuration, the average latency is approximately 299.5 ms per image, yielding a throughput of 3.34 FPS (see Table~\ref{tab:profiling_unified}).

\paragraph{Theoretical Complexity.}
Because the inference pipeline relies on a continuous-time RK4 ODE solver and \texttt{torch.autograd.grad} for energy scoring, standard automatic profiling tools (e.g., \texttt{fvcore} or \texttt{thop}) cannot reliably trace the computational graph to measure FLOPs. Instead, the computational complexity can be characterized theoretically as:
\begin{equation}
\mathcal{O}\Big(N_{\text{tokens}} \times \big(N_{\text{steps}} \cdot C_{\text{flow}} + C_{\text{ebm}}\big)\Big)
\label{eq:app_flops}
\end{equation}
where $N_{\text{tokens}}$ represents the number of spatial patch tokens extracted by the foundation backbone, $N_{\text{steps}}$ is the number of numerical solver steps (10 steps for RK4), and $C_{\text{flow}}$ and $C_{\text{ebm}}$ denote the cost of one flow-network evaluation and one EBM scoring operation, respectively.

\begin{table}[ht]
\centering
\caption{Inference profiling metrics for the Unified ReFP-AD pipeline. Measured on a single NVIDIA H100 GPU using cached DINOv2-G/14 features.}
\label{tab:profiling_unified}
\begin{tabular}{l c}
\toprule
\textbf{Metric} & \textbf{Value} \\
\midrule
Trainable Parameters & 13.12 M \\
~~-- \textit{Trainable Backbone Parameters} & \textit{0.00 M} \\
~~-- \textit{Rectified Flow} & \textit{9.45 M} \\
~~-- \textit{EBM} & \textit{3.67 M} \\
\midrule
Peak VRAM & 5198.5 MB \\
Average Latency & 299.5 ms \\
Throughput & 3.34 FPS \\
\bottomrule
\end{tabular}
\end{table}

\section{Input Resolution Ablation}
\label{sec:app_resolution_ablation}

We also evaluate the sensitivity of the original ReFP-AD configuration to input resolution on MVTec-AD. The backbone and model configuration are kept fixed, while the input resolution is varied across $224{\times}224$, $336{\times}336$, and $448{\times}448$. This isolates whether the proposed method relies on high-resolution inputs for detection and localization.

As shown in Table~\ref{tab:app_resolution_ablation}, ReFP-AD is already highly effective at $224{\times}224$, achieving 98.2\% Image AUROC and 97.9\% Pixel AUROC. Increasing the resolution to $448{\times}448$ improves Image AUROC to 98.6\%, while Pixel AUROC remains stable around 97.9\%. The gains are therefore modest but consistent, suggesting that higher resolution mainly benefits image-level discrimination, whereas token-level localization remains robust across resolutions.

\begin{table}[ht]
\centering
\caption{
Input resolution ablation for the original ReFP-AD configuration on MVTec-AD.
Performance is reported as Image AUROC / Pixel AUROC (\%).
}
\label{tab:app_resolution_ablation}
\setlength{\tabcolsep}{12pt}
\renewcommand{\arraystretch}{1.12}
\begin{tabular}{l cc}
\toprule
Input resolution & Image AUROC & Pixel AUROC \\
\midrule
$224{\times}224$ & 98.2 & 97.9 \\
$336{\times}336$ & 98.3 & 97.9 \\
$448{\times}448$ & \textbf{98.6} & \textbf{97.9} \\
\bottomrule
\end{tabular}
\end{table}

\section{Generalization to DINOv3-7B}
\label{sec:app_dinov3_backbone}

To further test whether ReFP-AD depends on a specific self-supervised backbone, we replace the DINOv2-G/14 backbone with the DINOv3-7B variant. This substantially increases the token dimensionality from 1536 to 4096. The remaining pipeline is kept unchanged, including unified rectified-flow preconditioning, unified EBM training, and gradient-norm scoring.

Table~\ref{tab:app_dinov3_backbone} reports the resulting performance. ReFP-AD remains stable with 4096-dimensional DINOv3-7B features, achieving 98.2\% / 97.9\% Image/Pixel AUROC on MVTec-AD and 96.8\% / 98.9\% on VisA. Compared with the main DINOv2-G/14 configuration, DINOv3-7B yields comparable localization performance, while image-level detection is slightly lower on VisA. This result indicates that the proposed flow-preconditioned EBM is not tied to DINOv2 features and remains effective even when the feature dimensionality is substantially increased.

\begin{table}[ht]
\centering
\caption{
Backbone generalization under the unified ReFP-AD protocol.
DINOv3-7B increases the token dimensionality from 1536 to 4096.
Performance is reported as Image AUROC / Pixel AUROC (\%).
}
\label{tab:app_dinov3_backbone}
\setlength{\tabcolsep}{12pt}
\renewcommand{\arraystretch}{1.12}
\begin{tabular}{l c c c}
\toprule
Backbone & Feature dim. & MVTec-AD & VisA \\
\midrule
DINOv2-G/14 & 1536 & 98.6 / 97.9 & 97.3 / 99.0 \\
DINOv3-7B  & 4096 & 98.2 / 97.9 & 96.8 / 98.9 \\
\bottomrule
\end{tabular}
\end{table}

\section{Qualitative Analysis and Localization}
\label{sec:app_qual}

\paragraph{Extended Visualizations.}
This section presents an extended qualitative comparison of the proposed method, ReFP-AD, against prior state-of-the-art approaches. The comparison includes the energy-based model MPDR~\cite{yoon2023energy} and the hierarchical normalizing flow method HGAD~\cite{yao2024hierarchical}. We assess the localization capabilities across multiple challenging defect categories sampled from both the MVTec-AD~\cite{bergmann2019mvtec} and VisA~\cite{zou2022spot} datasets.

\begin{figure}[ht!]
 \centering
 \caption{Extended qualitative localization results. Comparison of anomaly heatmaps generated by MPDR$^\dagger$~\cite{yoon2023energy}, HGAD~\cite{yao2024hierarchical}, and our \emph{ReFP-AD} on the MVTec-AD and VisA datasets, evaluated against the GT annotations.}
 \includegraphics[width=\textwidth]{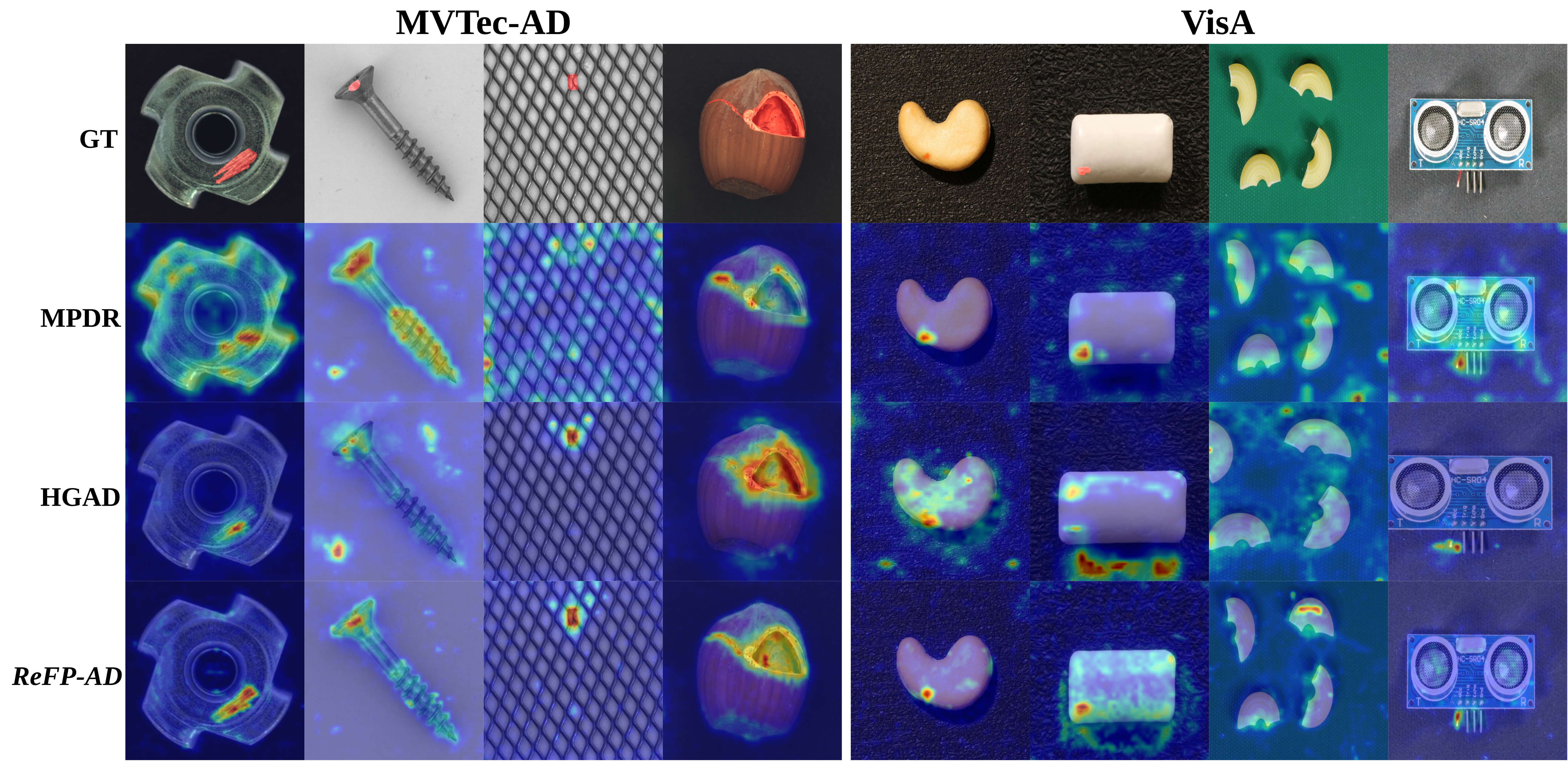}
 \label{fig:suppl_qual_extended}
\end{figure}

\paragraph{Observations.}
Figure~\ref{fig:suppl_qual_extended} illustrates the predicted anomaly heatmaps alongside the corresponding GT segmentation masks. While previous density-based methods often produce diffuse or noisy activation maps in highly heterogeneous unified settings, ReFP-AD generates consistently sharp and well-defined defect localizations. The generated energy gradients closely map to the precise boundaries of the GT regions, even for fine-grained anomalies. These qualitative results further support the effectiveness of the proposed geometric preconditioning framework, which preserves the meaningful spatial and semantic token relationships necessary for accurate anomaly detection.

\section{Outlook}
The supplementary results suggest several directions for future work. First, the strong performance of flow-native scoring indicates that geometric transport itself carries a substantial anomaly signal, motivating more principled objectives that align transport learning more directly with downstream density separation. Second, the checkpoint-selection analysis reveals that useful transports are characterized by their impact on short-run MCMC rather than regression loss alone, suggesting that sampler-aware training criteria may further improve robustness. Finally, extending the proposed geometric preconditioning framework to higher-resolution, multimodal, or temporally structured inspection data represents a promising direction for scalable anomaly detection beyond current image-based benchmarks.


\end{document}